\documentclass{article}

\PassOptionsToPackage{numbers, sort&compress}{natbib}

\usepackage[preprint]{neurips_2025}

\usepackage[utf8]{inputenc} %
\usepackage[T1]{fontenc}    %
\usepackage[
    colorlinks=true,   %
    linkcolor=black,    %
    citecolor=black,   %
    urlcolor=black   %
]{hyperref}
\usepackage{url}            %
\usepackage{booktabs}       %
\usepackage{amsfonts}       %
\usepackage{nicefrac}       %
\usepackage{microtype}      %
\usepackage{xcolor}         %
\usepackage{graphicx}
\usepackage{subcaption}
\usepackage{amsmath}
\usepackage{xcolor} %
\usepackage{colortbl} %
\definecolor{Gray}{gray}{0.92}
\usepackage{multirow}
\usepackage{comment}
\usepackage{wrapfig}
\usepackage{caption} 
\usepackage{amssymb}
\usepackage{enumitem}
\usepackage{mathrsfs}

\newtheorem{definition}{Definition}

\title{Global-Recent Semantic Reasoning on Dynamic Text-Attributed Graphs with Large Language Models}

\newcommand{\MethodName}{DyGRASP}

\author{%
Yunan Wang \quad \quad \quad \quad Jianxin Li \quad \quad \quad \quad Ziwei Zhang\thanks{Corresponding Author} \\
  School of Computer Science and Engineering \\ 
  Beihang University \\
}

\begin{document}

\maketitle

\begin{abstract}
Dynamic Text-Attribute Graphs (DyTAGs), characterized by time-evolving graph interactions and associated text attributes, are prevalent in real-world applications. Existing methods, such as Graph Neural Networks (GNNs) and Large Language Models (LLMs), mostly focus on static TAGs. Extending these existing methods to DyTAGs is challenging as they largely neglect the \textit{recent-global temporal semantics}: the recent semantic dependencies among interaction texts and the global semantic evolution of nodes over time. Furthermore, applying LLMs to the abundant and evolving text in DyTAGs faces efficiency issues. To tackle these challenges, we propose \underline{Dy}namic \underline{G}lobal-\underline{R}ecent \underline{A}daptive \underline{S}emantic \underline{P}rocessing (\MethodName), a novel method that leverages LLMs and temporal GNNs to efficiently and effectively reason on DyTAGs. Specifically, we first design a node-centric implicit reasoning method together with a sliding window mechanism to efficiently capture recent temporal semantics. In addition, to capture global semantic dynamics of nodes, we leverage explicit reasoning with tailored prompts and an RNN-like chain structure to infer long-term semantics. Lastly, we intricately integrate the recent and global temporal semantics as well as the dynamic graph structural information using updating and merging layers. Extensive experiments on DyTAG benchmarks demonstrate \MethodName's superiority, achieving up to 34\% improvement in Hit@10 for destination node retrieval task. Besides, \MethodName~exhibits strong generalization across different temporal GNNs and LLMs.
\end{abstract}

\section{Introduction}\label{sec:intro}
\begin{figure}[!t]
\centering
\includegraphics[width=1.0\columnwidth]{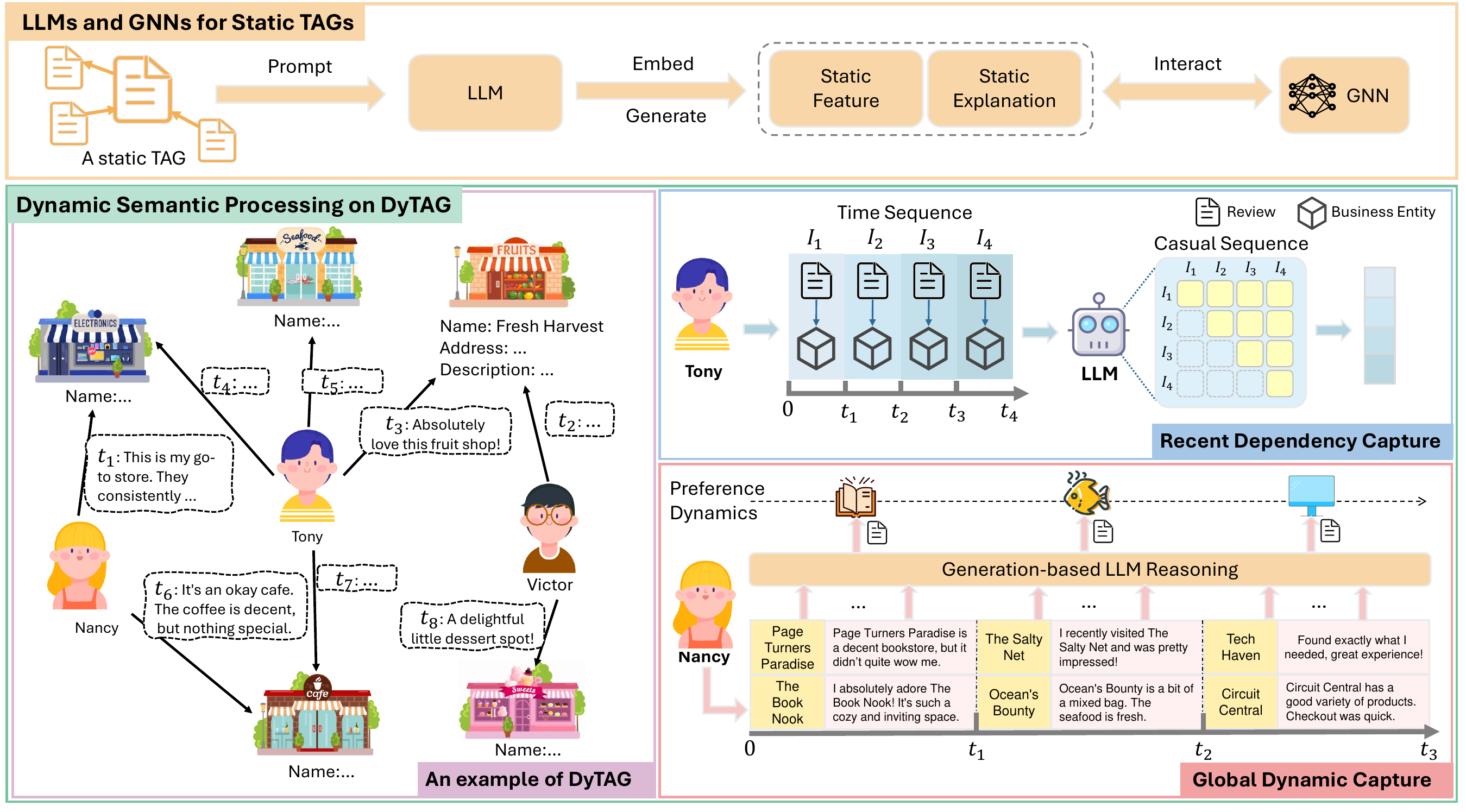}
\caption{An illustrative comparison between handling static TAGs and DyTAGs. Static methods that utilize LLMs usually generate static embeddings or explanations for text attributes. In comparison, DyTAGs contain rich spatio-temporal information, such as recent semantic dependency or global semantic dynamics of nodes, which the existing methods for TAGs do not account for.}
\label{fig:intro}
\vspace{-0.5cm}
\end{figure}

Dynamic Text-Attribute Graphs (DyTAGs) are widely present in real-world scenarios like E-commerce, knowledge graphs, and social networks \citep{khrabrov2010discovering,deng2019learning,song2019session,zhang2022dynamic,tang2023dynamic,luo2023hope,huang2022ttergm}. Unlike commonly studied TAGs~\citep{sen2008collective,giles1998citeseer,mernyei2020wiki,hu2020open,he2023explanations,yan2023comprehensive}, where nodes only contain static text attributes, DyTAGs contain evolving information over time and involve interactions accompanied by timestamp and dynamic text attributes~\citep{zhangdtgb}. For example, consider an E-comment graph illustrated in Figure \ref{fig:intro} where nodes represent users or merchants and interactions represent user reviews to merchants. Clearly, the spatio-temporal patterns of DyTAGs contain more abundant information than static TAGs. How to fully mine the rich value underlying DyTAGs is essential for both academia and industry.

For static TAGs, most primitive approaches resort to Graph Neural Networks (GNNs) \citep{wu2019simplifying,li2021training,velivckovic2018graph,hamilton2017inductive,kipf2017semi} for their end-to-end learning capabilities. While excelling at capturing graph structural information, these methods usually only adopt shallow text encodings, such as Bag-of-Words or word embeddings as features, thus lacking strong semantic knowledge to understand the textual attributes comprehensively. Recently, Large Language Models (LLMs) \citep{abdin2024phi,almazrouei2023falcon,touvron2023llama,guo2025deepseek,achiam2023gpt}, pretrained on vast text corpora, have demonstrated strong capabilities in text understanding and generation. Thus, researchers have explored various methods to combine GNNs and LLMs for TAGs~\citep{zhu2025graphclip,wang2024bridging,khoshraftar2025graphit,beiranvand2025integrating,chencurriculum}, aiming to integrate textual information into graph structures. Despite the remarkable progress in combining LLMs and GNNs for TAGs, these methods cannot be directly transferred to DyTAGs due to the following challenges:

Firstly, existing GNNs and LLMs neglect the global-recent semantic dynamics in DyTAGs. In real life, the temporal changes of many phenomena conform to a complicated mixture of recent and global patterns. For example, ice cream sales might be related to recent promotional activities while also being influenced by long-term dietary structure changes. Specifically, we investigate two complementary temporal semantic features with different temporal granularities within DyTAG: 
\vspace{-0.25cm}
\begin{itemize}[leftmargin=15pt, itemsep=0pt]
\item \textbf{Recent semantic dependency of temporal interactions:} The text attribute of an interaction in DyTAG exhibits recent semantic dependency. For instance, the word ``notebook'' might mean ``notepad'' following a ``visit bookstore'' interaction, but the same word is more likely to mean ``computer'' after a ``visit electronics store'' interaction. 
\item \textbf{Global semantic dynamics of nodes:} Unlike static node semantics, node features in DyTAG also continuously undergo global changes, which are reflected by emerging interactions. For example, a user in an E-commerce network may shift her interests from literature to technology, as reflected in visiting different kinds of shops. %
\end{itemize}
\vspace{-0.25cm}
Current GNNs and LLMs methods for TAGs largely ignore the above issues as they only employ static methods to model the semantic and structural information. 
. %

Secondly, DyTAGs pose more severe efficiency challenges for LLMs. For static TAGs, most text attributes are associated with nodes. However, for DyTAGs, there exist edge-level text (such as the example in Figure~\ref{fig:intro}) and the edge attributes can also evolve with time. Considering that the number of edges is usually orders of magnitude larger than the number of nodes and there can exist at least hundreds or thousands of time stamps, developing effective models for DyTAGs demands highly efficient LLM reasoning designs to reduce computational costs.

To tackle these challenges, in this paper, we propose \MethodName (Dynamic Global-Recent Adaptive Semantic Processing), which leverages both the implicit and the explicit reasoning capabilities of LLM to capture recent-global temporal semantics in DyTAGs. Specifically, \MethodName~features three novel designs: (i) Implicit reasoning for recent temporal semantics: to capture recent semantic dependency while retaining a low computational complexity, %
we leverage the unidirectional consistency between temporal sequences and causal sequences of LLMs to achieve node-centric implicit reasoning. %
In addition, we propose a sliding window mechanism to organize a node's historical interactions chronologically in batches for the LLM. Our proposed method models recent semantic features for multiple interactions simultaneously while ensuring no future information leakage, which significantly reduces the number of consumed tokens so that our method can be scaled to real-world DyTAGs; (ii) Explicit reasoning for global temporal semantics: to learn node representations from extensive historical interactions while reducing consumed number of tokens, we utilize the generative reasoning ability of LLM and segment a node's historical interactions into a constant number of partitions based on timestamps. Then, we carefully design prompts to instruct the LLM to summarize a node's feature description for each period. Finally, we employ an RNN-like reasoning chain structure to pass long-distance semantics and dynamically update node features; (iii) Integrating Semantics and Graph Structure: lastly, we design three tailored layers to update recent semantics, global semantics, and graph structural features and then integrate them, together with a temporal GNN to learn comprehensive node representations. %

To verify \MethodName's effectiveness, we conduct extensive experiments on DyTAGs benchmark. Our proposed method achieves up to a 34\% improvement in Hit@10 for destination node retrieval task compared to state-of-the-art methods. Besides, \MethodName~demonstrates strong generalization across different LLMs and Temporal GNNs. We also perform detailed analyses for different components and hyperparameters. Our contributions are summarized as follows:
\begin{itemize}[leftmargin=15pt, itemsep=0pt]
\item We study the recent-global spatio-temporal patterns in DyTAGs, which are overlooked in existing GNNs and LLMs for static TAGs.
\item We propose to leverage the implicit and explicit reasoning capabilities of LLMs to capture the recent-global semantics on DyTAGs. Motivated by this goal, we propose \MethodName, a tailored model fusing the advantages of LLMs and temporal GNNs.
\item We theoretically prove that our proposed method has optimized reasoning efficiency compared to straightforward methods, thereby reducing the cost of practical applications. 
\item We verify the effectiveness of \MethodName~through extensive experiments on DyTAG benchmarks, outperforming state-of-the-art baselines up to 34\%. 
\end{itemize}

\section{Related work}\label{sec:relatedwork}
\textbf{LLM for TAGs.} While GNNs excel at encoding graph structure features, preliminary methods exhibit deficiencies in understanding the textual content within TAGs. Consequently, researchers leverage the powerful capability of LLMs on text understanding to enhance the performance of GNNs on TAGs \citep{chiennode,yang2021graphformers,xue2023efficient,wu2024exploring,wei2024llmrec,guo2024graphedit,liuone}. Among them, ENGINE \citep{zhu2024efficient} employs LLMs to enhance the quality of node representations by proposing ``G-Ladder'' structure. TAPE \citep{he2024harnessing} utilizes ``explanations'' generated by an LLM for the textual attributes as features. SimTeG \citep{duan2023simteg} boosts textual graph learning by first fine-tuning a language model to generate node embeddings, which are then used as features for a separate GNN. GraphGPT \citep{tang2024graphgpt} employs graph instruction tuning to integrate LLMs with graph structural knowledge. However, the aforementioned methods primarily target TAGs containing only static semantic features. The dynamic characteristics of DyTAGs render these approaches incapable of modeling the temporal semantic relationships therein. In contrast, we investigate the temporal semantics inherent in DyTAGs by leveraging the reasoning capabilities of LLMs. %

\textbf{Temporal Graph Neural Networks}. To extend the capabilities of GNNs to dynamic graphs, recent studies have proposed various temporal GNNs \citep{rossi2020temporal,poursafaei2022towards,wang2021apan,luo2022neighborhood,ma2020streaming,cong2023we} to model dynamic graph features. Among these, TGAT \citep{xu2020inductive} introduces functional time encoding based on Bochner's theorem. CAWN \citep{wang2021inductive} leverages temporal random walks to inductively learn representations of temporal network dynamics. DyRep \citep{trivedi2019dyrep} learns dynamic graph representations by modeling a latent mediation process between topological evolution and node activities. DyGFormer~\citep{yu2023towards} learns solely from the historical first-hop interaction sequences of nodes to simplify the dynamic graph learning task. However, the aforementioned temporal GNNs are unable to comprehend the textual semantic information within DyTAG, leading to suboptimal performance. In comparison, our method introduces LLMs reasoning over DyTAG to capture the temporal semantic features.

\textbf{LLM for DyTAGs.} Following the pioneering benchmark for DyTAGs named DGTB~\citep{zhangdtgb}, there has recently been few research on LLMs for DyTAGs, which are largely concurrent to our work. Among them, LKD4DyTAG \citep{roy2025llm} distills knowledge from LLM into a temporal GNN. CROSS \citep{zhang2025unifying} employs LLMs to dynamically extract text semantics and a co-encoder to synergistically unify these semantics with evolving graph structures. GAD \citep{lei2025exploring} utilizes a multi-agent system with collaborative LLMs to directly perform prediction on DyTAGs without dataset-specific training. However, existing studies are unable to identify the recent-global patterns inherent in DyTAG, thus failing to comprehensively capture its multi-granularity temporal semantic features.

\section{Notations}\label{sec:notations}
\begin{definition}[DyTAG]
A DyTAG is represented as $\mathcal{G}=\{\mathcal{V},\mathcal{E},\mathcal{T},\mathcal{D},\mathcal{R}\}$, where $\mathcal{V}$ is the set of nodes, $\mathcal{E} \subseteq \mathcal{V} \times \mathcal{V}$ is the set of edges, $\mathcal{T}$ is the set of observed timestamps, $\mathcal{D}$ is the set of node textual attributes, and $\mathcal{R}$ is the set of edge textual attributes. The textual attribute of a node $u$ is denoted by $D_u \in \mathcal{D}$. An edge, also known as an interaction, occurring between nodes $u$ and $v$ at timestamp $t$ is represented as $I = (u,\mathbf{r},v,t)$, where $\mathbf{r} \in \mathcal{R}$ represents the textual attribute. 

\end{definition}
We use interaction to refer to an edge in DyTAGs to emphasize its dynamic nature, e.g., interaction text denotes the complete description of the source node's behavior towards the target node, rather than merely the edge's text attribute itself. We use reasoning to represent the LLM's process of uncovering underlying information from a series of interactions. We aim to develop models that can handle various tasks associated with DyTAGs, such as node classification, link prediction, and destination node retrieval. %

\section{Method}\label{sec:method}
In this section, we introduce our method. We first introduce implicit reasoning for recent temporal semantics in Section~\ref{sec:implicit reasoning}, then introduce explicit reasoning for global temporal semantics in Section \ref{sec:explicit reasoning}, and finally integrate the temporal semantics with the dynamic graph structures in Section \ref{sec:model}.

\subsection{Implicit Reasoning for Recent Temporal Semantics}\label{sec:implicit reasoning}
Unlike the static attributes in normal TAGs, the textual attributes in DyTAG contain temporal semantic relationships with historical interactions. Therefore, we leverage the implicit reasoning ability of LLMs by harvesting the hidden embedding of texts within LLMs.

To obtain textual attribute embeddings that incorporate temporal semantics, an intuitive approach is to take an edge-centric view and organize the textual attributes of an interaction and historical interactions chronologically, and use an LLM to extract the hidden embeddings for the interaction. %
However, denoting $d$ as the average degree of the graph, the complexity (i.e., the number of consumed input tokens) of the LLM reaches $O(|\mathcal{E}| \times d)$. 
For a real-world DyTAG that has millions of edges and an average degree of hundreds or thousands, this straightforward method suffers from an excessively high computational complexity. 
To address this potential issue while preserving temporal semantic relationships, we propose two key designs: (i) node-centric implicit reasoning, which leverages the unidirectional consistency between the causal order inherent in LLMs and the chronological order of the dynamic graph; (ii) sliding window mechanism, which segments interaction sequences into overlapping batches to balance computational consumption and semantic feature modeling capability. Using these two designs, our proposed method can theoretically reduce the computational complexity to $O(|\mathcal{E}|)$ (Proof is shown in Appendix \ref{sec:proof}). Next, we elaborate on these two designs.

\begin{figure}[!t]
\centering
\includegraphics[width=1.0\columnwidth]{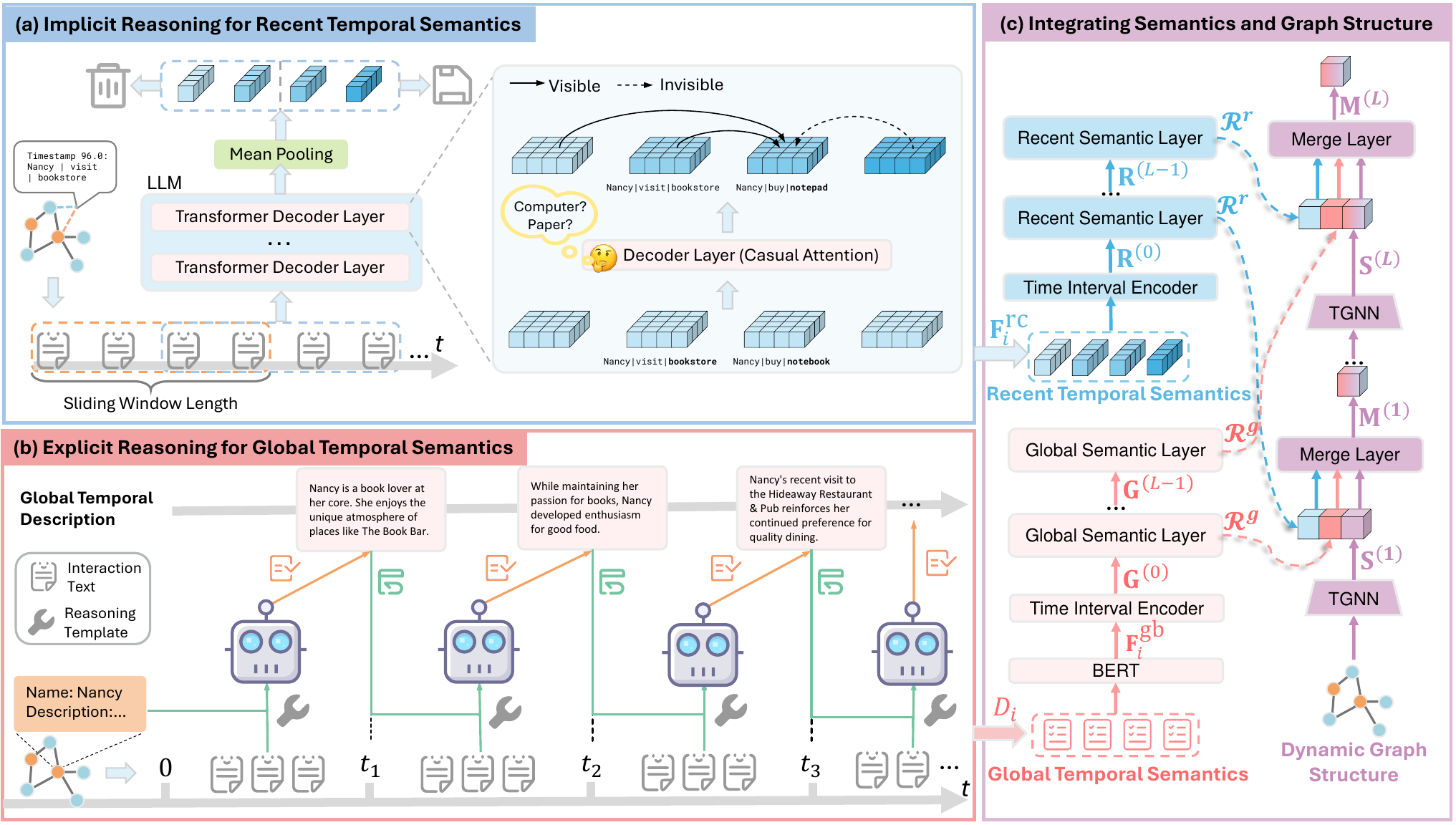}
\caption{An overview of \MethodName. \MethodName~first utilizes the implicit and explicit reasoning capabilities of the LLM to respectively extract recent and global temporal semantic features from a DyTAG, as shown in (a) and (b), respectively. Then it integrates these temporal semantic features with the dynamic graph structure features by temporal GNN, as shown in (c).}
\label{fig:method}
\vspace{-0.3cm}
\end{figure}

\textbf{Node-centric implicit reasoning.} As illustrated in Figure \ref{fig:method}(a), for a node $v$ in $\mathcal{G}$, we first extract all its interactions with its neighbors, denoted as $\mathcal{N}_v$. Then we sort them by the timestamp and arrange them into a sequence, which ensures interactions earlier in time positioned earlier in the sequence. Lastly, we input $\mathcal{N}_v$ into an LLM at once to extract hidden features of $\mathcal{N}_v$. As LLMs are designed for causal modeling of text sequences and share the unidirectional nature with the time sequence of $\mathcal{N}_v$, this consistency prevents an interaction from observing future interaction text, thereby avoiding leakage of future information. Benefiting from this, in a single forward pass, all interactions in $\mathcal{N}_v$ can simultaneously acquire recent semantic features from the historical interactions thus enabling efficient reasoning.

\textbf{Sliding window mechanism.} This module aims to solve three challenges: (i) the context window length limitation of LLMs, (ii) the heavy computational resource consumption, (iii) the tendency for LLM performance to degrade with increased input length \citep{wang2024beyond,levy2024same}. Specifically, we partition $\mathcal{N}_v$ into multiple overlapping batches:
\begin{equation}
\mathcal{B}_i = \{I_k | \frac{c}{2}*i +1\le k \le \frac{c}{2}*i + c, I_k \in \mathcal{N}_v\},
\end{equation}
where $\mathcal{B}_i$ is the $i$-th batch and $c$ is the window length. In each batch, the first $\frac{c}{2}$ interactions serve to provide temporal context for the subsequent $\frac{c}{2}$ interactions. Subsequently, for each batch, we organize the interactions chronologically into an input sequence using a dataset-specific prompt template (see an example in Figure \ref{fig:recent_prompt}) and feed it into the LLM. Finally, we apply mean pooling to the output hidden layer features for each interaction. %
The obtained temporal semantic feature for an interaction $i \in I$ is denoted as $\mathbf{F}^{\text{rc}}_i \in \mathbb{R}^{d_{\text{LLM}}}$, where $d_{\text{LLM}}$ denotes the dimensionality of the hidden features.

\subsection{Explicit Reasoning for Global Temporal Semantics}
\label{sec:explicit reasoning}
Complementary to recent temporal features, the occurrence of a real-world phenomenon is often associated with long-range global temporal features. Therefore, this module aims to uncover these global temporal semantic features with the explicit reasoning ability of LLMs, i.e., we directly let LLMs generate textual information after comprehending DyTAGs. 

Specifically, as shown in Figure \ref{fig:method}(b), we first evenly partition the interaction sequence $\mathcal{N}_v$ into $s$ segments by timestamps. Each segment is denoted by $S_i(1 \le i \le s)$. The partition timestamps are denoted by $\mathbf{T}^{\text{gb}}=\{\hat{t}_0, \hat{t}_1, \hat{t}_2, ..., \hat{t}_s\}$, where $\hat{t}_0=-1$ denotes the special starting time. $S_i$ consists of the chronologically ordered sequence of all interactions occurring between $\hat{t}_{i-1}$ (exclusive) and $\hat{t}_{i}$ (inclusive). Our motivation is twofold: (i) similar to implicit reasoning,  $\mathcal{N}_v$ cannot be fed into LLM at once; (ii) performing only a single inference fails to capture the temporal evolution of global semantic features.

Next, we employ an RNN-like reasoning chain structure to empower the LLM with memory capabilities. Specifically, let $D_i(1 \le i \le s)$ represent the textual description of the node's dynamic features corresponding to the period of $S_i$, and $D_0$ be the node's original textual attribute. The reasoning process can be described as:
\begin{equation}
    D_i = \text{LLM}\left(D_{i-1}, S_i\right),
\end{equation}
where we use dataset-specific template (see Figure \ref{fig:global_prompt} for an example) to prompt LLM to uncover the potential global semantics. The reasoning chain structure is designed for the propagation of long-range dynamic features and providing essential background description of the node for the LLM to infer the node's current features.

In the above process, each interaction is fed into the LLM twice. Consequently, the overall complexity is limited to  $O(|\mathcal{E}|)$, highlighting the efficiency of the explicit reasoning process.

\subsection{Integrating Semantics and Graph Structure}
\label{sec:model}
Next we present the component to integrate temporal semantics and dynamic graph structure features. For a node $v$ and a prediction time $t$, this module aims to obtain an integrated representation, which will be used for downstream tasks. As shown in Figure \ref{fig:method}(c), we first divide the model into three components: Recent Semantic (RS) layer,  Global Semantic (GS) layer, and graph structure layer. Let $L$ denote the total number of layers and $l$ denote the current layer index.

\textbf{Recent Semantic Layer.} Let $\mathcal{N}^t_v=\{I_i|t_i<t, I_i\in \mathcal{N}_v\}$. In Section~\ref{sec:implicit reasoning}, we have obtained the temporal feature $\mathbf{F}^{\text{rc}}_i$ for interaction $I_i \in \mathcal{N}^t_v$ and its corresponding timestamp $t_i$. Following \citep{xu2020inductive}, we use a learnable time encoder to encode the time interval $\Delta t_i=t-t_i$ into a $d_t$-dimensional vector to capture periodic temporal patterns:
\begin{equation}
    \mathbf{R}^{(0)}_i=\left[ \mathcal{P}\left(\mathbf{F}^{\text{rc}}_i\right) \ ||\ \mathscr{T}(t-t_i)\right] \label{eq:rf},
\end{equation}
where $\mathcal{P}(\cdot): \mathbb{R}^{d_{\text{LLM}}}\rightarrow \mathbb{R}^{d_t}$ is a projector, $\left[\cdot || \cdot \right]$ denotes concatenation, and $\mathscr{T}(\cdot)$ is the time encoder.
Note that $d_t \ll d_{\text{LLM}}$ and thus the projection can also accelerate computing. Lastly, $\{\mathbf{R}^{(0)}_i\}$ are used as the input feature for the RS layer to compute the recent temporal semantics:
\begin{equation}\label{eq:rs}
\mathbf{R}^{(l)} = \left[ \mathbf{R}^{(l)}_0,\mathbf{R}^{(l)}_1,...,\mathbf{R}^{(l)}_{|\mathcal{N}^t_v|} \right]=\text{RS\_Layer} \left(\left[\mathbf{R}^{(l-1)}_0,\mathbf{R}^{(l-1)}_1,...,\mathbf{R}^{(l-1)}_{|\mathcal{N}^t_v|}\right] \right). 
\end{equation}
Given the strengths of Transformers in processing sequential data, we employ a Transformer Encoder to realize $\text{RS\_Layer}(\cdot)$.

\textbf{Global Semantic Layer.} Similarly, we have obtained the global textual description $D_i$ for node $v$ in timestamp $\hat{t}_i$ in Section~\ref{sec:explicit reasoning}. We use a word embedding model (e.g., BERT) to transform each textual description $D_i$ into a hidden representation $\mathbf{F}^{\text{gb}}_i \in \mathbb{R}^{d_{\text{BERT}}}$ as global semantic features. To capture the dynamic changes in node features leading up to $t$, we then calculate the input features $\mathbf{G}^{(0)}_i$ as follows:
\begin{equation}
    \mathbf{G}^{(0)}_i= \left[ \mathcal{P}^\prime(\mathbf{F}^{\text{gb}}_i)\ ||\ \mathscr{T}(t-\hat{t}_i) \right],
\end{equation}
where $\mathcal{P}^\prime(\cdot): \mathbb{R}^{d_{\text{Bert}}}\rightarrow \mathbb{R}^{d_T}$ is another projector. Lastly, the global feature are updated as:
\begin{equation}
    \mathbf{G}^{(l)}= \left[\mathbf{G}^{(l)}_0,\mathbf{G}^{(l)}_1,...,\mathbf{G}^{(l)}_{\hat{i}}\right]=\text{GS\_Layer}\left(\left[\mathbf{G}^{(l-1)}_0,\mathbf{G}^{(l-1)}_1,...,\mathbf{G}^{(l-1)}_{\hat{i}}\right]\right).
\end{equation}
where $\hat{i}=\max\{i | \hat{t}_i < t\}$ ensures that only features up to $t$ are considered to prevent future information leakage. Similar to the RS\_layer, we use a Transformer Encoder to realize $\text{GS\_Layer}(\cdot)$ in our implementation.

\textbf{Graph Structure Layer.} In this component, we first explicitly model dynamic graph structures by learning the structural representation $\mathbf{S}^{(l)}$. The graph structure layer is represented as:
\begin{equation}
    \mathbf{S}^{(l)}=\text{TGNN}\left(\mathbf{M}^{(l-1)}, \text{MP}_{\mathcal{G}}(v,t)\right),
\end{equation}
where $\text{MP}_{\mathcal{G}}(v,t)$ denotes the process of gathering and aggregating features from the neighbors of node $v$ in graph $\mathcal{G}$ up to time $t$, $\text{TGNN}(\cdot)$ is a temporal GNN, and $\mathbf{M}^{(l-1)}$ is the integrated node feature. Benefiting from the modularity design of \MethodName, our method is compatible with any existing message-passing temporal GNN. In our implementation, two representative temporal GNNs, DyGFormer \citep{yu2023towards} and TGAT \citep{xu2020inductive}, are used to demonstrate the compatibility of \MethodName. 

The integrated node feature $\mathbf{M}^{(l)}$ aims to fuse features from three components, i.e., recent semantics $\{\mathbf{R}^{(l)}_i\}$, global semantics $\{\mathbf{G}^{(l)}_i\}$), and graph structure $\mathbf{S}^{(l)}$, using $\text{Merge\_Layer}(\cdot)$:
\begin{equation}
    \mathbf{M}^{(l)}=\text{Merge\_Layer}\left(\mathcal{R}^r(\{\mathbf{R}^{(l)}_i\}), \mathcal{R}^g(\{\mathbf{G}^{(l)}_i\}), \mathbf{S}^{(l)}\right), 
\end{equation}
where $\mathbf{M}^{(l)} \in \mathbb{R}^{d_{\text{SF}}}$ represents the node's integrated feature, $\mathbf{M}^{(0)}$ is the node raw feature, $\mathcal{R}^{r}(\cdot)$ and $\mathcal{R}^{g}(\cdot)$ are readout functions responsible for summarizing the features from the RS and GS Layer, respectively. We set $\text{Merge\_Layer}$ as an MLP and  $\mathcal{R}^r(\cdot)$ as a mean pooling operation:
\begin{equation}
    \mathcal{R}^r(\{\mathbf{R}^{(l)}_i\}) = \text{Mean\_Pooling}\left([\mathbf{R}^{(l)}_0, \mathbf{R}^{(l)}_1, \ldots ]\right).
\end{equation}
For $\mathcal{R}^g(\cdot)$, we simply take the last feature in the sequence, i.e., 
\begin{equation}
    \mathcal{R}^g(\{\mathbf{G}^{(l)}_i\}) = \mathbf{G}^{(l)}_{\hat{i}},
\end{equation}
which is based on following considerations: (i) the Transformer Encoder in the $\text{GS\_Layer}(\cdot)$ inherently aggregates information via the attention mechanism, potentially concentrating historical global semantics in the final feature $\mathbf{G}^{(l)}_{\hat{i}}$; (ii) the preceding global reasoning chain employs a RNN-like mechanism to propagate historical global semantic to the final feature.  The final output of the model, $\mathbf{M}^{(L)}$, serves as the node's comprehensive feature representation for downstream tasks.

\section{Experiments}\label{sec:exp}
In this section, we first show the superiority of \MethodName~over existing methods across various tasks and validate the generalizability of \MethodName~on multiple LLMs and base temporal GNNs. Then we separately validate the effectiveness of Recent/Global Semantic Reasoning and study the impact of the hyperparameter, as well as investigate the inference efficiency.

\subsection{Experimental settings}
 We select four datasets from the DTGB benchmark~\citep{zhangdtgb}: GDELT, Enron, Googlemap, and Stack\_elec, which cover diverse graph scales and domains. We choose seven state-of-the-art temporal GNNs as baselines: JODIE \citep{kumar2019predicting}, DyRep \citep{trivedi2019dyrep}, CAWN \citep{wang2021inductive}, TCL \citep{wang2021tcl}, GraphMixer \citep{cong2023we}, TGAT \citep{xu2020inductive}, and DyGFormer \citep{yu2023towards}. We also evaluate the performance of Llama-3.1-8B-Instruct \citep{grattafiori2024llama}, the primary LLM used in \MethodName, as another baseline. Following \citep{zhangdtgb}, we evaluate models using the destination node retrieval task with Hit@10 metric under transductive and inductive settings and the future link prediction task with AP and ROC-AUC metrics. More details can be found in Appendix~\ref{sec:detail_setting}.

\subsection{Main Results}\label{sec:main_results}
\begin{table*}[t]
    \centering
    \caption{Hit@10 (\%) results for destination node retrieval. OOM means out of memory. We use \textbf{boldface} and \underline{underlining} to denote the best and the second-best performance, respectively.}
    \resizebox{\textwidth}{!}{%
    \begin{tabular}{l|cccc|cccc}
    \toprule
      \multirow{2}{*}{\textbf{Methods} } & \multicolumn{4}{c|}{\textbf{Transductive}}  & \multicolumn{4}{c}{\textbf{Inductive}} \\
      &\textbf{GDELT}&\textbf{Enron}&\textbf{Googlemap}&\textbf{Stack\_elec}&\textbf{GDELT}&\textbf{Enron}&\textbf{Googlemap}&\textbf{Stack\_elec} \\
     \midrule 
      \textbf{JODIE} & 89.88\scalebox{0.75}{±0.10} & 91.01\scalebox{0.75}{±0.82} & OOM & OOM & 68.84\scalebox{0.75}{±1.09} & 78.78\scalebox{0.75}{±1.63} & OOM & OOM \\
      \textbf{DyRep} & 85.34\scalebox{0.75}{±0.61} & 85.79\scalebox{0.75}{±2.81} & OOM & OOM & 61.63\scalebox{0.75}{±4.01} & 68.53\scalebox{0.75}{±1.68} & OOM & OOM \\
      \textbf{CAWN} & 89.92\scalebox{0.75}{±0.24} & 91.96\scalebox{0.75}{±0.87} & 52.07\scalebox{0.75}{±0.97} & 91.37\scalebox{0.75}{±0.17} & 63.84\scalebox{0.75}{±0.99} & 76.48\scalebox{0.75}{±1.03} & 44.04\scalebox{0.75}{±0.74} & 46.30\scalebox{0.75}{±0.32} \\
      \textbf{TCL} & 90.58\scalebox{0.75}{±0.44} & 87.96\scalebox{0.75}{±3.12} & 48.70\scalebox{0.75}{±0.75} & 84.16\scalebox{0.75}{±10.94} & 69.28\scalebox{0.75}{±1.12} & 64.99\scalebox{0.75}{±9.99} & 40.90\scalebox{0.75}{±0.66} & 37.82\scalebox{0.75}{±11.58} \\
      \textbf{GraphMixer} & 88.67\scalebox{0.75}{±0.05} & 87.36\scalebox{0.75}{±1.07} & 48.34\scalebox{0.75}{±0.23} & 93.25\scalebox{0.75}{±0.04} & 63.92\scalebox{0.75}{±0.12} & 66.83\scalebox{0.75}{±1.08} & 40.71\scalebox{0.75}{±0.19} & 52.04\scalebox{0.75}{±0.34}\\
     \cmidrule{1-9}
      \textbf{Llama-3.1} & 41.01 & 92.09 & 53.97 & 34.24 & 31.75 & 79.55 & 43.56 & 33.59 \\
     \cmidrule{1-9}
      \textbf{TGAT} & 89.97\scalebox{0.75}{±0.03} & 89.12\scalebox{0.75}{±0.21} & 63.53\scalebox{0.75}{±0.69} & 93.74\scalebox{0.75}{±0.11} & 65.76\scalebox{0.75}{±0.46} & 70.93\scalebox{0.75}{±0.85} & 56.61\scalebox{0.75}{±0.51} & 54.29\scalebox{0.75}{±0.47} \\
    \rowcolor{Gray}
     \textbf{\MethodName}\scalebox{0.75}{(TGAT)} & \underline{91.96\scalebox{0.75}{±0.17}} & \underline{96.25\scalebox{0.75}{±1.11}} & \underline{82.79\scalebox{0.75}{±0.17}} & \textbf{99.74\scalebox{0.75}{±0.04}} & 71.78\scalebox{0.75}{±0.28} & \underline{86.76\scalebox{0.75}{±2.07}} & \underline{78.63\scalebox{0.75}{±0.20}} & \textbf{99.77\scalebox{0.75}{±0.09}} \\
      \rowcolor{Gray}
     \textbf{\textit{Improv.}} $\uparrow$ & \textit{+1.99} & \textit{+7.13} & \textit{+19.26} & \textit{+6.00} & \textit{+6.02} & \textit{+15.83} & \textit{+22.02} & \textit{+45.48}\\
    \cmidrule{1-9}
     \textbf{DyGFormer} & 91.64\scalebox{0.75}{±0.24} & 92.04±1.7 & 51.32\scalebox{0.75}{±2.13} & 94.39\scalebox{0.75}{±0.12} & \underline{72.49\scalebox{0.75}{±0.37}} & 82.52\scalebox{0.75}{±2.34} & 43.84\scalebox{0.75}{±1.92} & 56.23\scalebox{0.75}{±0.52} \\
    \rowcolor{Gray}
     \textbf{\MethodName}\scalebox{0.75}{(DyGFormer)} & \textbf{93.24\scalebox{0.75}{±0.13}} & \textbf{99.40\scalebox{0.75}{±0.29}} & \textbf{85.88\scalebox{0.75}{±2.64}} & \underline{99.59\scalebox{0.75}{±0.25}} & \textbf{75.93\scalebox{0.75}{±0.16}} & \textbf{95.18\scalebox{0.75}{±0.58}} & \textbf{81.14\scalebox{0.75}{±2.12}} & \underline{99.74\scalebox{0.75}{±0.14}}\\
    \rowcolor{Gray}
     \textbf{\textit{Improv.}} $\uparrow$
 & \textit{+1.60} & \textit{+7.36} & \textit{+34.56} & \textit{+5.20} & \textit{+3.44} & \textit{+12.66} & \textit{+37.30} & \textit{+43.51}\\
    \bottomrule
    \end{tabular}}
    \label{tab:main}
    \vspace{-0.3cm}
\end{table*}

\begin{figure}[t]
    \centering
    \begin{subfigure}[b]{0.49\textwidth}
        \centering
        \includegraphics[width=\textwidth]{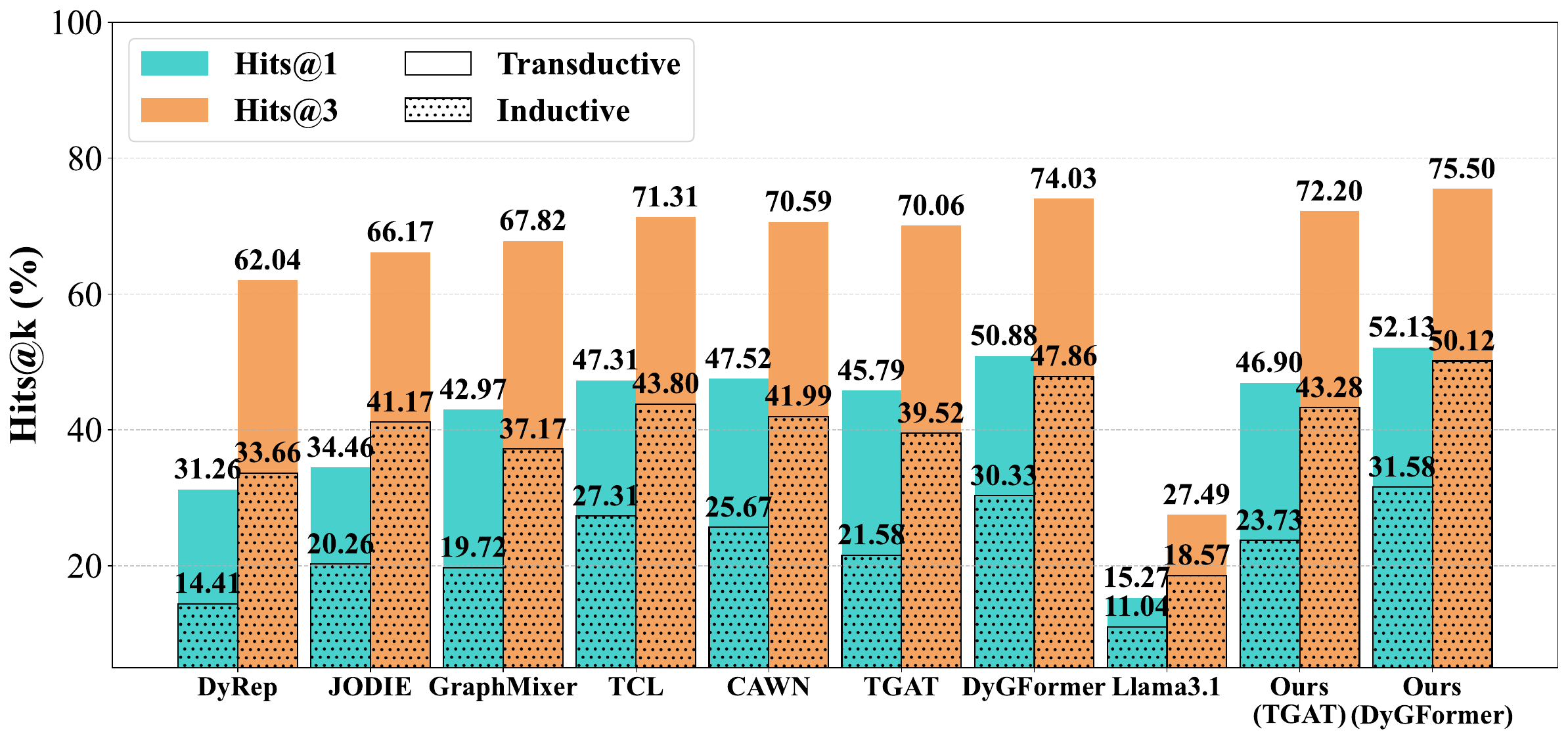}
        \vspace{-0.6cm}
        \caption{GDELT}
        \label{fig:sub1}
    \end{subfigure}
    \hfill
    \begin{subfigure}[b]{0.49\textwidth}
        \centering
        \includegraphics[width=\textwidth]{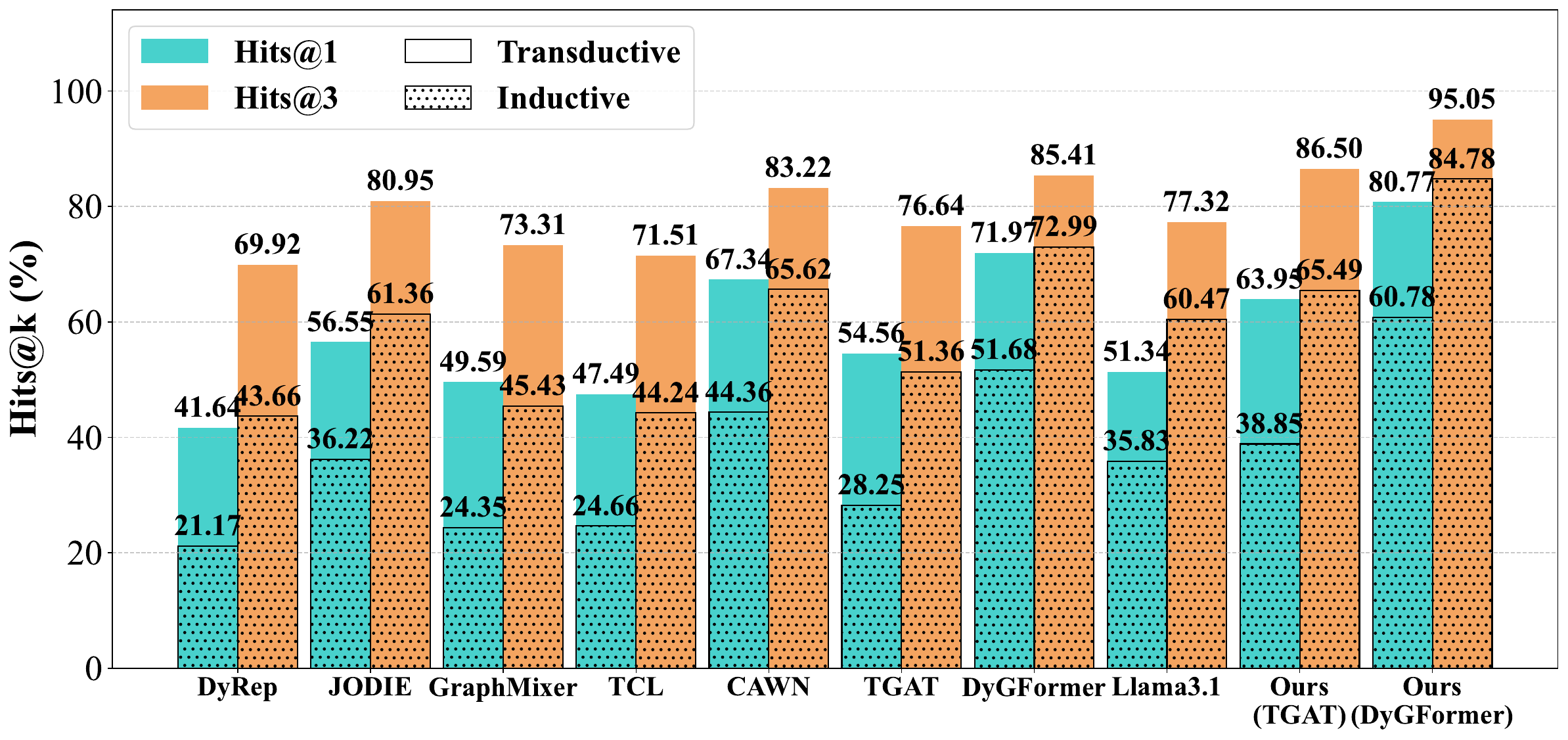}
         \vspace{-0.6cm}
        \caption{Enron}
        \label{fig:sub2}
    \end{subfigure}
    \begin{subfigure}[b]{0.49\textwidth}
        \centering
        \includegraphics[width=\textwidth]{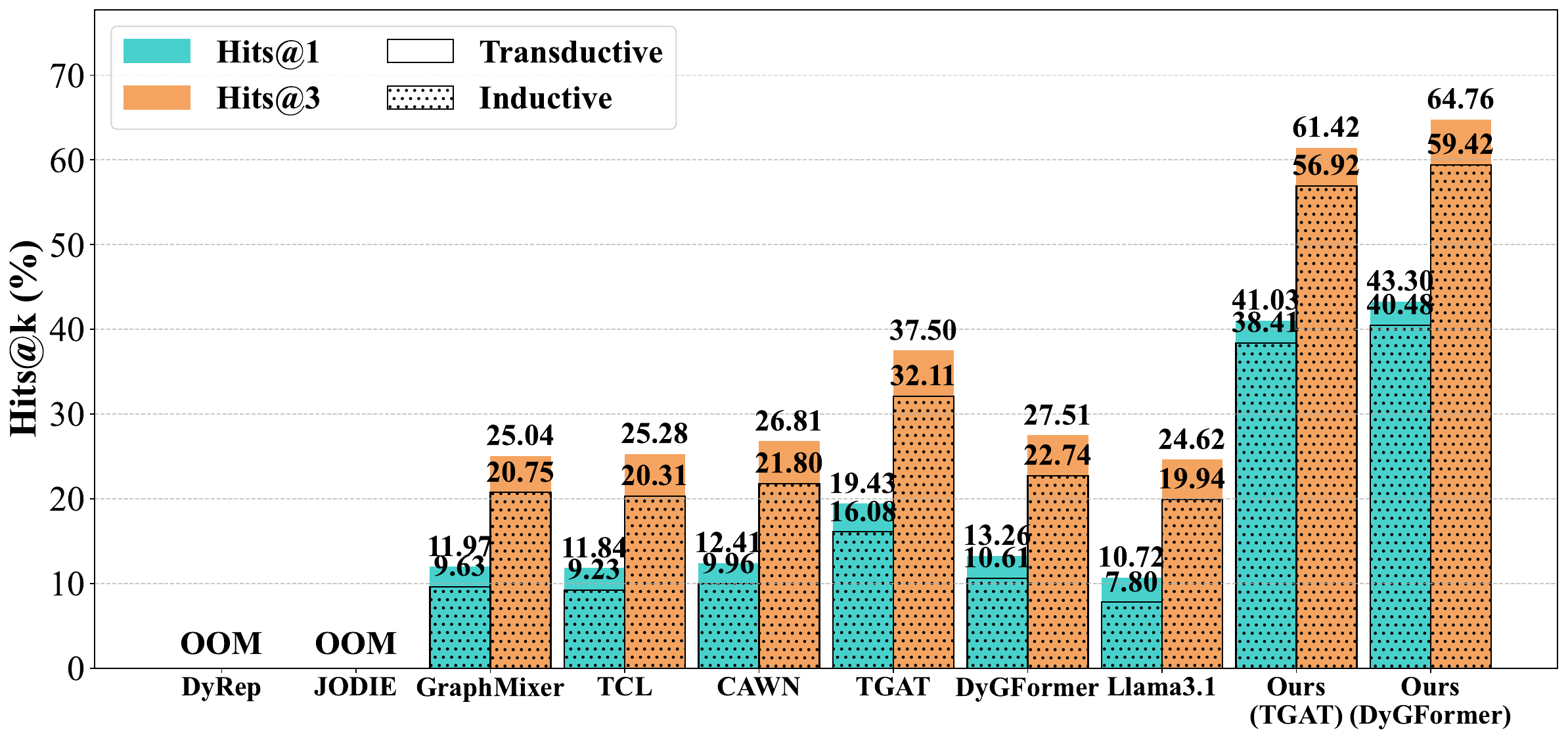}
         \vspace{-0.6cm}
        \caption{Googlemap}
        \label{fig:sub3}
    \end{subfigure}
    \hfill
    \begin{subfigure}[b]{0.49\textwidth}
            \centering
            \includegraphics[width=\textwidth]{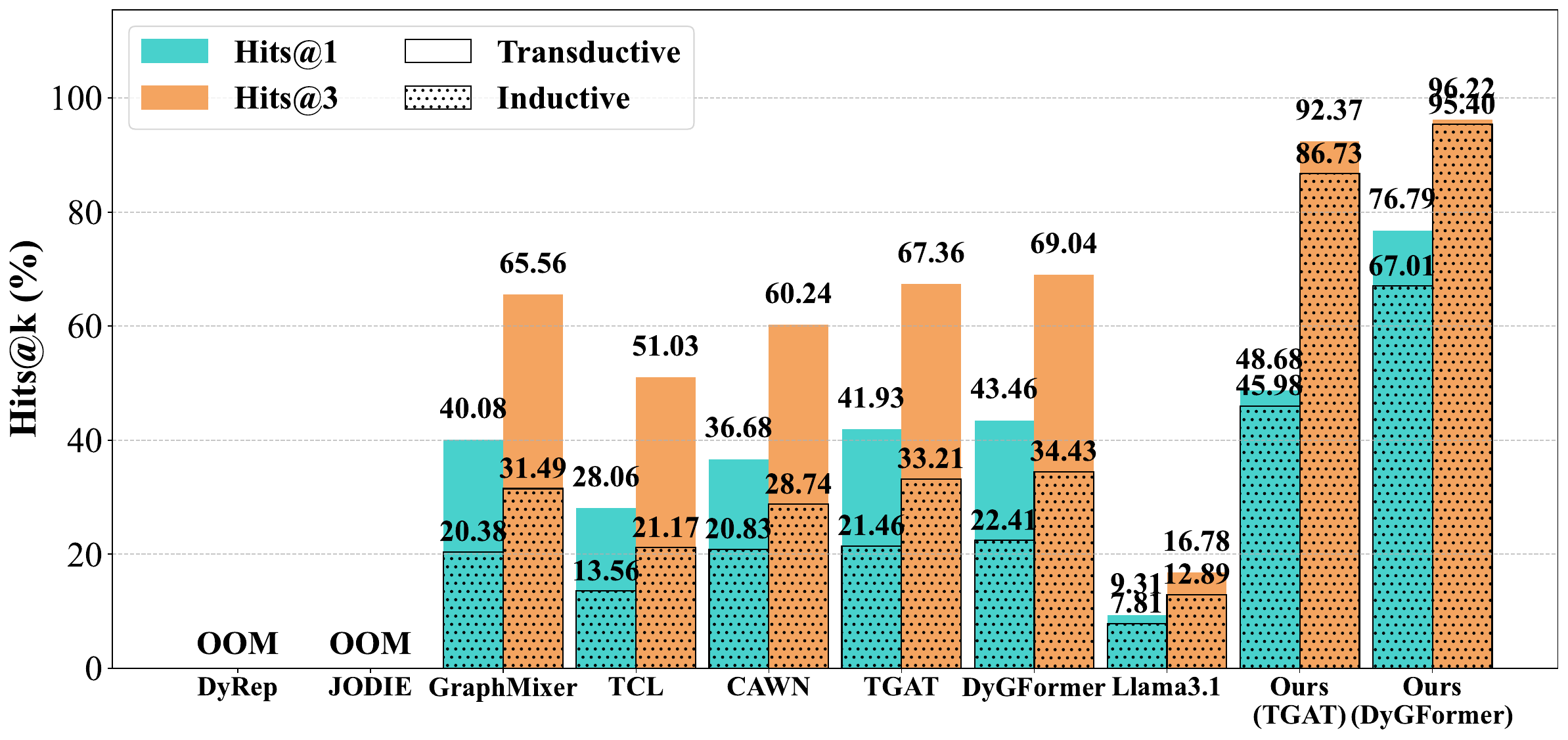}
             \vspace{-0.6cm}
            \caption{Stack\_elec}
            \label{fig:sub4}
        \end{subfigure}
    \caption{Hit@1/3 (\%) results for destination node retrieval.}
    \label{fig:main_figure}
\vspace{-0.3cm}
\end{figure}

\begin{table*}[htp]
    \centering
    \caption{Results for future link prediction. OOM means out of memory. We use \textbf{boldface} and \underline{underlining} to denote the best and the second-best performance, respectively.}
    \vspace{-5pt}
    \resizebox{\textwidth}{!}{%
    \begin{tabular}{p{0.85cm}|l|cccc|cccc}
    \toprule
     \multirow{2}{*}{\textbf{Metric} }& \multirow{2}{*}{\textbf{Methods} } & \multicolumn{4}{c|}{\textbf{Transductive}}  & \multicolumn{4}{c}{\textbf{Inductive}} \\
      & &\textbf{GDELT}&\textbf{Enron}&\textbf{Googlemap}&\textbf{Stack\_elec}&\textbf{GDELT}&\textbf{Enron}&\textbf{Googlemap}&\textbf{Stack\_elec} \\
     \midrule 
    \multirow{9}{*}{\textbf{\textit{AP}}} 
     & \textbf{JODIE} & 95.56\scalebox{0.75}{±0.08} & 96.70\scalebox{0.75}{±0.34} & OOM & OOM & 87.94\scalebox{0.75}{±0.47} & 90.26\scalebox{0.75}{±0.58} & OOM & OOM \\
     & \textbf{DyRep} & 94.55\scalebox{0.75}{±0.19} & 94.99\scalebox{0.75}{±0.93} & OOM & OOM & 85.02\scalebox{0.75}{±1.59} & 85.84\scalebox{0.75}{±1.11} & OOM & OOM \\
     & \textbf{CAWN} & 96.57\scalebox{0.75}{±0.05} & 97.85\scalebox{0.75}{±0.13} & 82.78\scalebox{0.75}{±0.50} & 95.55\scalebox{0.75}{±0.08} & 88.65\scalebox{0.75}{±0.29} & 92.91\scalebox{0.75}{±0.38} & 78.67\scalebox{0.75}{±0.63} & 80.48\scalebox{0.75}{±0.21} \\
     & \textbf{TCL} & 96.63\scalebox{0.75}{±0.14} & 96.31\scalebox{0.75}{±1.44} & 80.71\scalebox{0.75}{±0.40} & 89.58\scalebox{0.75}{±9.50} & 89.95\scalebox{0.75}{±0.38} & 88.27\scalebox{0.75}{±4.37} & 76.13\scalebox{0.75}{±0.46} & 73.09\scalebox{0.75}{±10.44} \\
     & \textbf{GraphMixer} & 95.95\scalebox{0.75}{±0.05} & 96.01\scalebox{0.75}{±0.22} & 80.60\scalebox{0.75}{±0.05} & 96.17\scalebox{0.75}{±0.02} & 87.30\scalebox{0.75}{±0.06} & 87.60\scalebox{0.75}{±0.30} & 76.13\scalebox{0.75}{±0.18} & 82.75\scalebox{0.75}{±0.20}\\
     & \textbf{TGAT} & 96.41\scalebox{0.75}{±0.02} & 96.78\scalebox{0.75}{±0.04} & 87.71\scalebox{0.75}{±0.27} & 96.59\scalebox{0.75}{±0.10} & 88.30\scalebox{0.75}{±0.14} & 89.54\scalebox{0.75}{±0.33} & 84.97\scalebox{0.75}{±0.23} & 84.24\scalebox{0.75}{±0.51} \\
     & \textbf{DyGFormer} & 97.02\scalebox{0.75}{±0.03} & 97.86\scalebox{0.75}{±0.01} & 81.63\scalebox{0.75}{±0.40} & 96.70\scalebox{0.75}{±0.10} & 91.09\scalebox{0.75}{±0.15} & \underline{93.60\scalebox{0.75}{±0.22}} & 77.23\scalebox{0.75}{±0.72} & 84.67\scalebox{0.75}{±0.34} \\
     \cmidrule{2-10}
     & \textbf{Llama-3.1} & 82.23 & 86.13 & 67.80 & 60.35 & 73.61 & 74.34 & 64.08 & 58.93 \\
     \cmidrule{2-10}
    \rowcolor{Gray}
    \multicolumn{1}{l|}{\cellcolor{white}}
    & \textbf{\MethodName}\scalebox{0.75}{(TGAT)}
 & \textbf{97.78\scalebox{0.75}{±1.46}} & \underline{98.04\scalebox{0.75}{±0.61}} & \textbf{96.23\scalebox{0.75}{±2.46}} & \underline{98.25\scalebox{0.75}{±0.13}} & \textbf{93.05\scalebox{0.75}{±4.43}} & 91.66\scalebox{0.75}{±3.26} & \textbf{95.31\scalebox{0.75}{±3.01}} & \underline{88.11\scalebox{0.75}{±0.31}} \\
    \rowcolor{Gray}
    \multicolumn{1}{l|}{\cellcolor{white}} 
    & \textbf{\MethodName}\scalebox{0.75}{(DyGFormer)} & \underline{97.40\scalebox{0.75}{±0.03}} & \textbf{98.81\scalebox{0.75}{±0.14}} & \underline{94.85\scalebox{0.75}{±0.22}} & \textbf{98.42\scalebox{0.75}{±0.02}} & \underline{92.23\scalebox{0.75}{±0.02}} & \textbf{95.03\scalebox{0.75}{±0.60}} & \underline{93.68\scalebox{0.75}{±0.20}} & \textbf{88.50\scalebox{0.75}{±0.20}}\\
    \midrule
    \multirow{9}{*}{\textbf{\textit{AUC}}} 
     & \textbf{JODIE} & 96.26\scalebox{0.75}{±0.02} & 97.01\scalebox{0.75}{±0.25} & OOM & OOM & 88.03\scalebox{0.75}{±0.27} & 90.21\scalebox{0.75}{±0.52} & OOM & OOM\\
     & \textbf{DyRep} & 95.23\scalebox{0.75}{±0.11} & 95.47\scalebox{0.75}{±0.92} & OOM & OOM & 85.49\scalebox{0.75}{±0.69} & 86.24\scalebox{0.75}{±1.03} & OOM & OOM\\
     & \textbf{CAWN} & 96.69\scalebox{0.75}{±0.05} & 97.80\scalebox{0.75}{±0.17} & 83.45\scalebox{0.75}{±0.50} &96.38\scalebox{0.75}{±0.10}& 88.32\scalebox{0.75}{±0.29} & 91.93\scalebox{0.75}{±0.57} & 78.92\scalebox{0.75}{±0.67} & 80.44\scalebox{0.75}{±0.28} \\
     & \textbf{TCL} & 96.77\scalebox{0.75}{±0.13} & 96.54\scalebox{0.75}{±1.24} & 80.90\scalebox{0.75}{±0.41} & 89.23\scalebox{0.75}{±11.74} & 89.56\scalebox{0.75}{±0.38} & 88.08\scalebox{0.75}{±4.15} & 75.45\scalebox{0.75}{±0.66} & 73.16\scalebox{0.75}{±10.59} \\
     & \textbf{GraphMixer} & 96.14\scalebox{0.75}{±0.02} & 96.13\scalebox{0.75}{±0.21} & 80.84\scalebox{0.75}{±0.07} & 96.86\scalebox{0.75}{±0.02} & 87.19\scalebox{0.75}{±0.13} & 87.33\scalebox{0.75}{±0.26} & 75.21\scalebox{0.75}{±0.25} & 83.25\scalebox{0.75}{±0.13}\\
     & \textbf{TGAT} & 96.58\scalebox{0.75}{±0.02} & 96.92\scalebox{0.75}{±0.04} & 88.19\scalebox{0.75}{±0.33} & 97.19\scalebox{0.75}{±0.06} & 88.33\scalebox{0.75}{±0.12} & 89.46\scalebox{0.75}{±0.34} & 85.34\scalebox{0.75}{±0.30} & 84.56\scalebox{0.75}{±0.29} \\
     & \textbf{DyGFormer} & 97.10\scalebox{0.75}{±0.03} & 97.69\scalebox{0.75}{±0.03} & 81.91\scalebox{0.75}{±0.37} & 97.31\scalebox{0.75}{±0.07} & 90.59\scalebox{0.75}{±0.15} & \underline{92.68\scalebox{0.75}{±0.28}} & 76.34\scalebox{0.75}{±0.78} & 85.24\scalebox{0.75}{±0.31} \\
     \cmidrule{2-10}
     & \textbf{Llama-3.1} & 83.31 & 86.86 & 69.38 & 55.41 & 74.03 & 78.26 & 66.17 & 56.04 \\
     \cmidrule{2-10}
    \rowcolor{Gray}
    \multicolumn{1}{l|}{\cellcolor{white}}
    & \textbf{\MethodName}\scalebox{0.75}{(TGAT)}
 & \underline{97.22\scalebox{0.75}{±0.04}} & \underline{98.34\scalebox{0.75}{±0.39}} & \underline{94.87\scalebox{0.75}{±0.04}} & \underline{98.79\scalebox{0.75}{±0.06}} & \underline{90.74\scalebox{0.75}{±0.10}} & 92.38\scalebox{0.75}{±2.33} & \underline{93.47\scalebox{0.75}{±0.08}} & \underline{85.91\scalebox{0.75}{±0.39}} \\
    \rowcolor{Gray}
    \multicolumn{1}{l|}{\cellcolor{white}} 
    & \textbf{\MethodName}\scalebox{0.75}{(DyGFormer)} & \textbf{97.60\scalebox{0.75}{±0.04}} & \textbf{98.86\scalebox{0.75}{±0.14}} & \textbf{94.89\scalebox{0.75}{±0.26}} & \textbf{98.87\scalebox{0.75}{±0.01}} & \textbf{92.19\scalebox{0.75}{±0.06}} & \textbf{94.85\scalebox{0.75}{±0.63}} & \textbf{93.58\scalebox{0.75}{±0.25}} & \textbf{86.69\scalebox{0.75}{±0.20}}\\
    \bottomrule
    \end{tabular}}
    \label{tab:future_link}
    \vspace{-0.3cm}
\end{table*}

\textbf{\MethodName~consistently outperforms all existing methods with substantial improvements.} As shown in Table \ref{tab:main} we can observe that under both transductive and inductive settings, \MethodName(DyGFormer) outperforms existing methods on all datasets, including traditional temporal GNNs and Llama-3.1-8B-Instruct, demonstrating the superiority of \MethodName. Notably, benefiting from the strong generalization of LLMs on semantic understanding, %
\MethodName~achieves higher improvements in the inductive setting compared to the transductive setting. Moreover, as shown in Table \ref{tab:main} and Figure \ref{fig:dataset_text}, the performance improvement over the base temporal GNN is more pronounced on datasets with richer textual information, highlighting the important role of the LLM in our method. 

\begin{figure}[t] 
\centering
\begin{minipage}[t]{0.48\linewidth} 
\centering 
    \includegraphics[width=1.0\columnwidth]{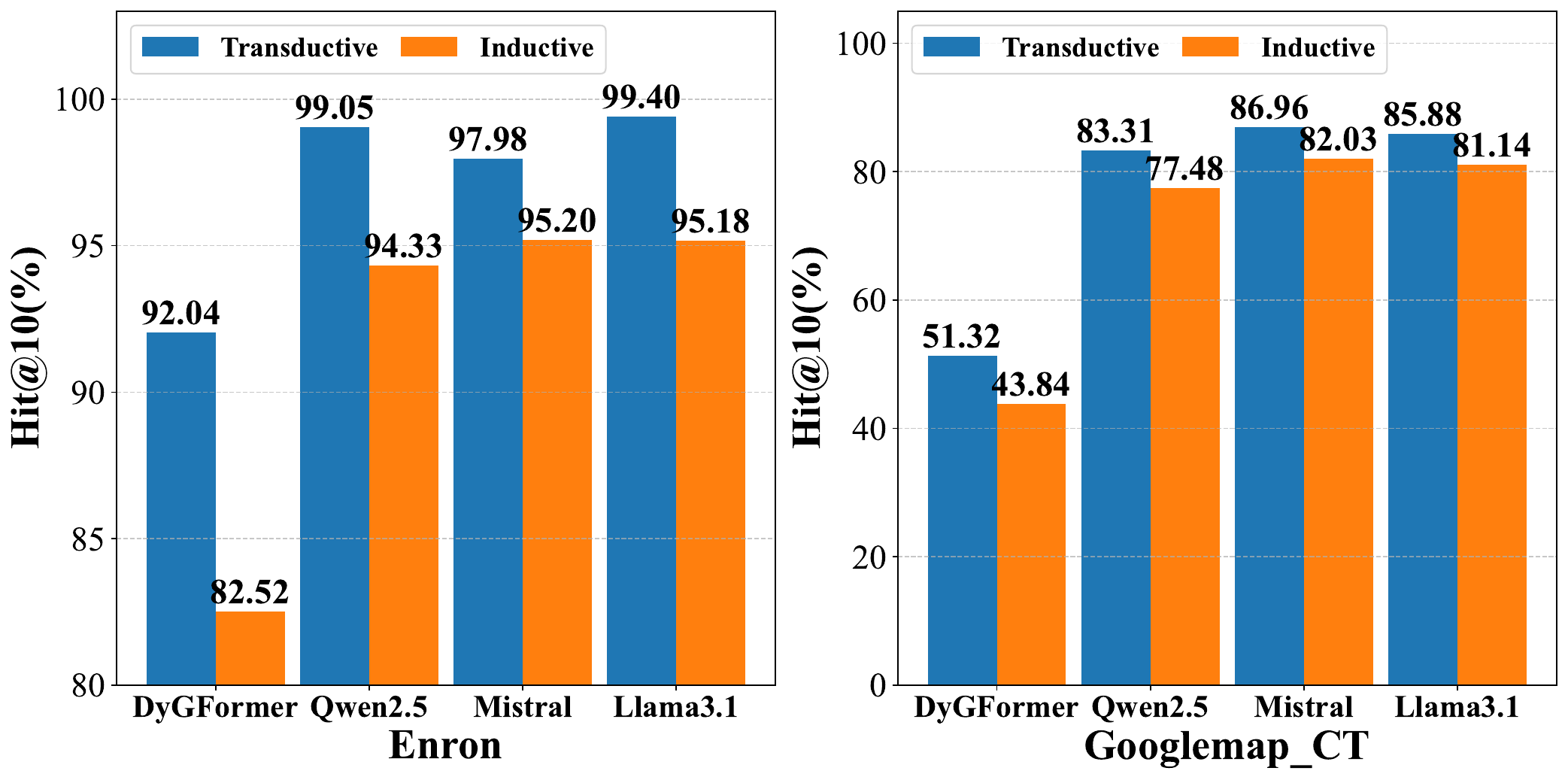}
    \vspace{-0.4cm}
    \caption{Results with different LLM backbones. Our method is compatible with various LLMs.}
    \label{fig:qwen_llama}
\end{minipage} 
\hfill
\begin{minipage}[t]{0.50\linewidth} 
\centering 
    \includegraphics[width=1.0\columnwidth]{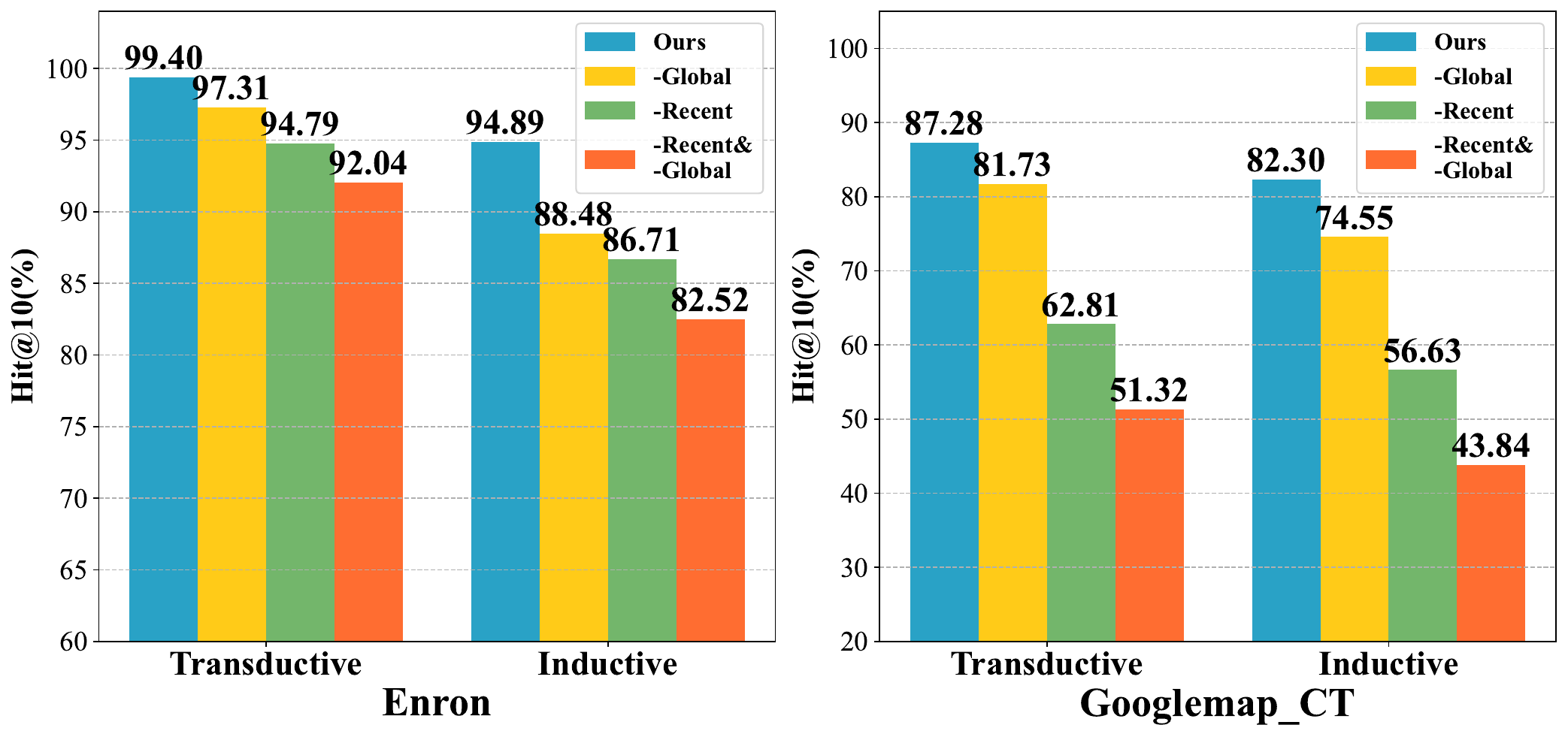}
    \vspace{-0.4cm}
    \caption{Results of ablation studies. Both global and recent modules contribute to our method. }
    \label{fig:ablation}
\end{minipage} 
 \vspace{-0.5cm}
\end{figure}

\textbf{\MethodName~demonstrates generalizability across both base temporal GNNs and LLMs.} As shown in Table \ref{tab:main} , the results indicate that \MethodName~consistently yields significant performance improvements compared to the original temporal GNNs when either using TGAT or DyGFormer as the base temporal GNN. Furthermore, we validate the effectiveness of \MethodName~with three different families of LLMs, including Qwen2.5, Mistral, and Llama-3.1, on the Enron and Googlemap datasets. As shown in Figure \ref{fig:qwen_llama}, \MethodName~consistently outperforms the base temporal GNN regardless of the specific LLM and exhibits minimal variation across different LLMs, demonstrating the exceptional general applicability and compatibility of \MethodName.

\subsection{Ablation Studies and Analyses}\label{sec:ablation}
\textbf{Ablation studies.}
To validate the components of our method, we evaluate the following ablations: (i) removing the global semantic layer (denoted as -Global); (ii) removing the recent semantic layer (denoted as -Recent); (iii) removing both layers (denoted as -Recent\&-Global).

As illustrated in Figure \ref{fig:ablation}, under both the transductive and inductive settings, adding either the recent or global temporal semantic reasoning module individually leads to improvements in performance. Furthermore, the performance obtained with both modules added simultaneously surpassed those achieved when only a single module is incorporated, which strongly confirms the complementarity of recent-global features and the effectiveness of both LLM reasoning designs.

\textbf{Hyperparameter Sensitivity Analysis}. Next, we investigate the key hyperparameters of \MethodName, including the number of global segments $s$ in global semantic reasoning and the maximum interaction length $c$ in recent temporal semantic reasoning. We separately study the effects of $c$ and $s$ using the features given by either the recent semantic layer or the global semantic layer. We report results on the Googlemap dataset, while other datasets show similar trends.

\begin{wrapfigure}{r}{0.6\textwidth}
    \centering
    \includegraphics[width=\linewidth]{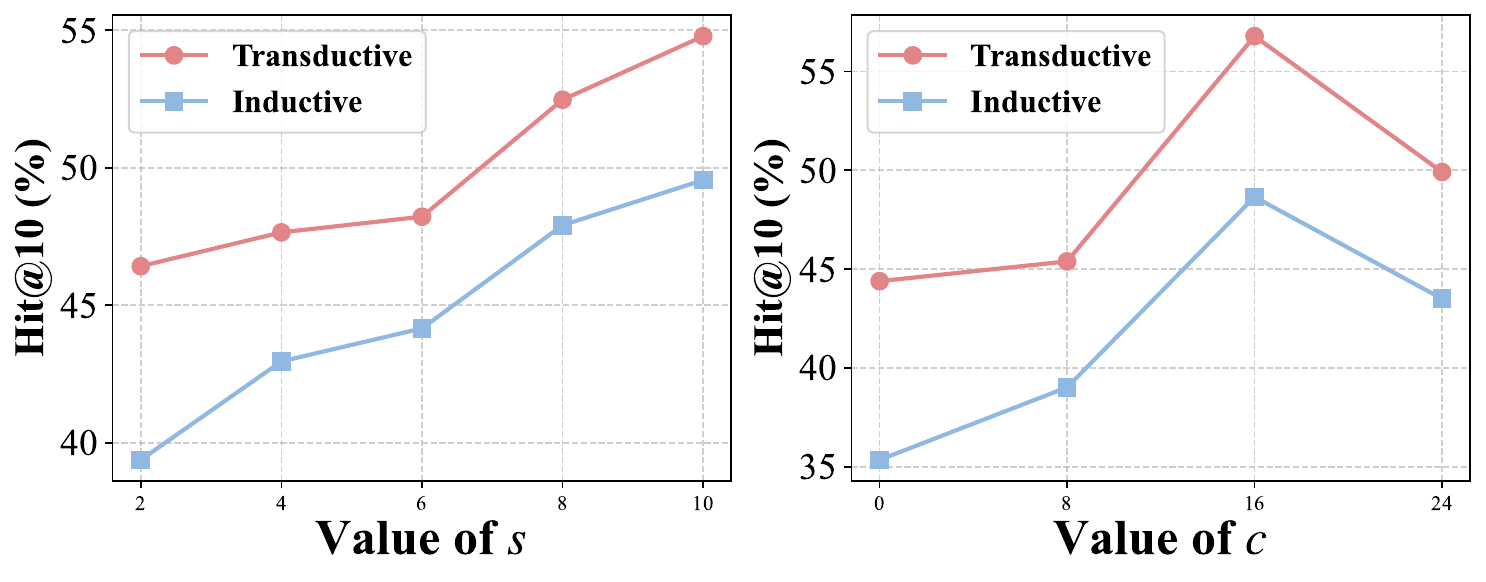}
    \caption{Results of sensitivity analysis.}
    \label{fig:sensitive}
\end{wrapfigure}

\textit{Number of global segments $s$}. As shown in Figure \ref{fig:sensitive}, model performance consistently improves as $s$ increases. We hypothesize that it is because a larger number of segments leads to more frequent reasoning for the LLM and smaller intervals between two reasoning time points, which allows the LLM to give smoother global temporal semantic variations. However, a higher number of $s$ also leads to a proportional increase in the inference time. %

\textit{Maximum interaction length $c$}. As illustrated in Figure \ref{fig:sensitive}, model performance initially improves significantly as $c$ increases but then decreases. A plausible reason is that, compared to independent encoding of text attribute ($c=0$), the recent semantic reasoning captures temporal semantic dependencies between interactions. A larger $c$ allows observation of more historical interactions, enriching the recent temporal semantics thus leading to performance gains. However, when $c$ surpasses 16, performance starts to decline plausibly because of increased input length, i.e., the performance of LLMs can degrade significantly as input length increases~\citep{wang2024beyond,levy2024same}. %

\textbf{Efficiency Analysis.}\label{sec:efficiency}
Due to the decoupling of LLM inference from the online prediction task, the inference of actual prediction task can be as fast as a lightweight temporal GNN. Furthermore, the computationally intensive LLM reasoning is an offline, one-time pre-processing step to extract temporal semantic features. These features are then stored (e.g., in a vector database) for fast retrieval. Here we investigate the inference efficiency of LLMs in our method. The results are shown in Table \ref{tab:reasoning_time} where the inference time is measured by Nvidia-A100 GPU-Days, 

\begin{table*}[h]
    \centering
    \caption{Token consumption and inference time of different components.}
    \resizebox{\textwidth}{!}{
    \begin{tabular}{l|cccc|cccc}
    \toprule
     \multirow{2}{*}{\textbf{Reasoning configuration} } & \multicolumn{4}{c|}{\textbf{Token Consumption}}  & \multicolumn{4}{c}{\textbf{Inference time}} \\
&\textbf{GDELT}&\textbf{Enron}&\textbf{Googlemap}&\textbf{Stack\_elec}&\textbf{GDELT}&\textbf{Enron}&\textbf{Googlemap}&\textbf{Stack\_elec} \\
     \midrule 
        Recent w/o node-centric reasoning & 1.68B & 3.62B & 2.42B & 7.00B & 2.37 & 4.34 & 2.45 & 7.21 \\
        Recent w/ node-centric reasoning & 0.10B & 0.86B & 0.35B & 0.96B & 0.16 & 0.84 & 0.52 & 1.21 \\
        \midrule
        Global Reasoning& 0.03B & 0.25B & 0.52B & 0.48B &0.04 & 0.37 & 0.89 & 1.42\\
        \bottomrule
    \end{tabular}}
    \label{tab:reasoning_time}
    \vspace{-0.3cm}
\end{table*}

\textit{Recent semantic reasoning efficiency.} As discussed in Section \ref{sec:implicit reasoning}, we introduce node-centric implicit reasoning to address the inference efficiency. The results show that the node-centric reasoning method significantly reduces the reasoning time across all datasets, which is consistent with our theoretical analysis.%

\textit{Global semantic reasoning efficiency.} We also report the inference time in Table \ref{tab:reasoning_time}. We could observe that the time consumption for this module is also low, highlighting the efficiency of \MethodName.

\section{Conclusion}\label{sec:conclusion}
In this paper, we propose \MethodName, an efficient and effective model for DyTAGs by capturing the recent-global temporal semantics. 
We leverage the implicit and explicit reasoning capabilities of LLMs to uncover the recent and global semantics in DyTAGs, respectively, and integrate these temporal semantic features with dynamic graph structural features through a temporal GNN. Experiments show that \MethodName~achieves significant improvement of up to 34\% in the Hit@10 metric for the destination node retrieval task compared to state-of-the-art methods. %

\bibliographystyle{plain}
\bibliography{reference.bib}

\begin{thebibliography}{10}

\bibitem{abdin2024phi}
Marah Abdin, Jyoti Aneja, Harkirat Behl, S{\'e}bastien Bubeck, Ronen Eldan, Suriya Gunasekar, Michael Harrison, Russell~J Hewett, Mojan Javaheripi, Piero Kauffmann, et~al.
\newblock Phi-4 technical report.
\newblock {\em arXiv preprint arXiv:2412.08905}, 2024.

\bibitem{achiam2023gpt}
Josh Achiam, Steven Adler, Sandhini Agarwal, Lama Ahmad, Ilge Akkaya, Florencia~Leoni Aleman, Diogo Almeida, Janko Altenschmidt, Sam Altman, Shyamal Anadkat, et~al.
\newblock Gpt-4 technical report.
\newblock {\em arXiv preprint arXiv:2303.08774}, 2023.

\bibitem{almazrouei2023falcon}
Ebtesam Almazrouei, Hamza Alobeidli, Abdulaziz Alshamsi, Alessandro Cappelli, Ruxandra Cojocaru, Merouane Debbah, Etienne Goffinet, Daniel Heslow, Julien Launay, Quentin Malartic, et~al.
\newblock Falcon-40b: an open large language model with state-of-the-art performance, 2023.

\bibitem{beiranvand2025integrating}
Azadeh Beiranvand and Seyed~Mehdi Vahidipour.
\newblock Integrating structural and semantic signals in text-attributed graphs with bigtex.
\newblock {\em arXiv preprint arXiv:2504.12474}, 2025.

\bibitem{chencurriculum}
Haibo Chen, Xin Wang, Zeyang Zhang, Haoyang Li, Weigao Wen, Ling Feng, and Wenwu Zhu.
\newblock Curriculum gnn-llm alignment for text-attributed graphs.
\newblock 2024.

\bibitem{chiennode}
Eli Chien, Wei-Cheng Chang, Cho-Jui Hsieh, Hsiang-Fu Yu, Jiong Zhang, Olgica Milenkovic, and Inderjit~S Dhillon.
\newblock Node feature extraction by self-supervised multi-scale neighborhood prediction.
\newblock In {\em International Conference on Learning Representations}.

\bibitem{cong2023we}
Weilin Cong, Si~Zhang, Jian Kang, Baichuan Yuan, Hao Wu, Xin Zhou, Hanghang Tong, and Mehrdad Mahdavi.
\newblock Do we really need complicated model architectures for temporal networks?
\newblock In {\em 11th International Conference on Learning Representations, ICLR 2023}, 2023.

\bibitem{deng2019learning}
Songgaojun Deng, Huzefa Rangwala, and Yue Ning.
\newblock Learning dynamic context graphs for predicting social events.
\newblock In {\em Proceedings of the 25th ACM SIGKDD international conference on knowledge discovery \& data mining}, pages 1007--1016, 2019.

\bibitem{devlin2019bert}
Jacob Devlin, Ming-Wei Chang, Kenton Lee, and Kristina Toutanova.
\newblock Bert: Pre-training of deep bidirectional transformers for language understanding.
\newblock In {\em Proceedings of the 2019 conference of the North American chapter of the association for computational linguistics: human language technologies, volume 1 (long and short papers)}, pages 4171--4186, 2019.

\bibitem{duan2023simteg}
Keyu Duan, Qian Liu, Tat-Seng Chua, Shuicheng Yan, Wei~Tsang Ooi, Qizhe Xie, and Junxian He.
\newblock Simteg: A frustratingly simple approach improves textual graph learning.
\newblock {\em arXiv preprint arXiv:2308.02565}, 2023.

\bibitem{giles1998citeseer}
C~Lee Giles, Kurt~D Bollacker, and Steve Lawrence.
\newblock Citeseer: An automatic citation indexing system.
\newblock In {\em Proceedings of the third ACM conference on Digital libraries}, pages 89--98, 1998.

\bibitem{grattafiori2024llama}
Aaron Grattafiori, Abhimanyu Dubey, Abhinav Jauhri, Abhinav Pandey, Abhishek Kadian, Ahmad Al-Dahle, Aiesha Letman, Akhil Mathur, Alan Schelten, Alex Vaughan, et~al.
\newblock The llama 3 herd of models.
\newblock {\em arXiv preprint arXiv:2407.21783}, 2024.

\bibitem{guo2025deepseek}
Daya Guo, Dejian Yang, Haowei Zhang, Junxiao Song, Ruoyu Zhang, Runxin Xu, Qihao Zhu, Shirong Ma, Peiyi Wang, Xiao Bi, et~al.
\newblock Deepseek-r1: Incentivizing reasoning capability in llms via reinforcement learning.
\newblock {\em arXiv preprint arXiv:2501.12948}, 2025.

\bibitem{guo2024graphedit}
Zirui Guo, Lianghao Xia, Yanhua Yu, Yuling Wang, Zixuan Yang, Wei Wei, Liang Pang, Tat-Seng Chua, and Chao Huang.
\newblock Graphedit: Large language models for graph structure learning.
\newblock {\em arXiv preprint arXiv:2402.15183}, 2024.

\bibitem{hamilton2017inductive}
Will Hamilton, Zhitao Ying, and Jure Leskovec.
\newblock Inductive representation learning on large graphs.
\newblock {\em Advances in neural information processing systems}, 30, 2017.

\bibitem{he2023explanations}
Xiaoxin He, Xavier Bresson, Thomas Laurent, Bryan Hooi, et~al.
\newblock Explanations as features: Llm-based features for text-attributed graphs.
\newblock {\em arXiv preprint arXiv:2305.19523}, 2(4):8, 2023.

\bibitem{he2024harnessing}
Xiaoxin He, Xavier Bresson, Thomas Laurent, Adam Perold, Yann LeCun, and Bryan Hooi.
\newblock Harnessing explanations: Llm-to-lm interpreter for enhanced text-attributed graph representation learning.
\newblock In {\em 12th International Conference on Learning Representations, ICLR 2024}, 2024.

\bibitem{hu2020open}
Weihua Hu, Matthias Fey, Marinka Zitnik, Yuxiao Dong, Hongyu Ren, Bowen Liu, Michele Catasta, and Jure Leskovec.
\newblock Open graph benchmark: Datasets for machine learning on graphs.
\newblock {\em Advances in neural information processing systems}, 33:22118--22133, 2020.

\bibitem{huang2022ttergm}
Yifan Huang, Clayton~Thomas Barham, Eric Page, and PK~Douglas.
\newblock Ttergm: Social theory-driven network simulation.
\newblock In {\em NeurIPS 2022 Temporal Graph Learning Workshop}, 2022.

\bibitem{DBLP:journals/corr/abs-2310-06825}
Albert~Q. Jiang, Alexandre Sablayrolles, Arthur Mensch, Chris Bamford, Devendra~Singh Chaplot, Diego de~Las~Casas, Florian Bressand, Gianna Lengyel, Guillaume Lample, Lucile Saulnier, L{\'{e}}lio~Renard Lavaud, Marie{-}Anne Lachaux, Pierre Stock, Teven~Le Scao, Thibaut Lavril, Thomas Wang, Timoth{\'{e}}e Lacroix, and William~El Sayed.
\newblock Mistral 7b.
\newblock {\em CoRR}, abs/2310.06825, 2023.

\bibitem{khoshraftar2025graphit}
Shima Khoshraftar, Niaz Abedini, and Amir Hajian.
\newblock Graphit: Efficient node classification on text-attributed graphs with prompt optimized llms.
\newblock {\em arXiv preprint arXiv:2502.10522}, 2025.

\bibitem{khrabrov2010discovering}
Alexy Khrabrov and George Cybenko.
\newblock Discovering influence in communication networks using dynamic graph analysis.
\newblock In {\em 2010 IEEE Second International Conference on Social Computing}, pages 288--294. IEEE, 2010.

\bibitem{kingma2014adam}
Diederik~P Kingma and Jimmy Ba.
\newblock Adam: A method for stochastic optimization.
\newblock {\em International Conference on Learning Representations}, 2014.

\bibitem{kipf2017semi}
Thomas~N Kipf and Max Welling.
\newblock Semi-supervised classification with graph convolutional networks.
\newblock In {\em International Conference on Learning Representations}, 2017.

\bibitem{kumar2019predicting}
Srijan Kumar, Xikun Zhang, and Jure Leskovec.
\newblock Predicting dynamic embedding trajectory in temporal interaction networks.
\newblock In {\em Proceedings of the 25th ACM SIGKDD international conference on knowledge discovery \& data mining}, pages 1269--1278, 2019.

\bibitem{lei2025exploring}
Runlin Lei, Jiarui Ji, Haipeng Ding, Lu~Yi, Zhewei Wei, Yongchao Liu, and Chuntao Hong.
\newblock Exploring the potential of large language models as predictors in dynamic text-attributed graphs.
\newblock {\em arXiv preprint arXiv:2503.03258}, 2025.

\bibitem{levy2024same}
Mosh Levy, Alon Jacoby, and Yoav Goldberg.
\newblock Same task, more tokens: the impact of input length on the reasoning performance of large language models.
\newblock In {\em Proceedings of the 62nd Annual Meeting of the Association for Computational Linguistics (Volume 1: Long Papers)}, pages 15339--15353, 2024.

\bibitem{li2021training}
Guohao Li, Matthias M{\"u}ller, Bernard Ghanem, and Vladlen Koltun.
\newblock Training graph neural networks with 1000 layers.
\newblock In {\em International conference on machine learning}, pages 6437--6449, 2021.

\bibitem{liuone}
Hao Liu, Jiarui Feng, Lecheng Kong, Ningyue Liang, Dacheng Tao, Yixin Chen, and Muhan Zhang.
\newblock One for all: Towards training one graph model for all classification tasks.
\newblock In {\em The Twelfth International Conference on Learning Representations}, 2024.

\bibitem{luo2023hope}
Xiao Luo, Jingyang Yuan, Zijie Huang, Huiyu Jiang, Yifang Qin, Wei Ju, Ming Zhang, and Yizhou Sun.
\newblock Hope: High-order graph ode for modeling interacting dynamics.
\newblock In {\em International conference on machine learning}, pages 23124--23139, 2023.

\bibitem{luo2022neighborhood}
Yuhong Luo and Pan Li.
\newblock Neighborhood-aware scalable temporal network representation learning.
\newblock In {\em Learning on Graphs Conference}, pages 1--1, 2022.

\bibitem{ma2020streaming}
Yao Ma, Ziyi Guo, Zhaocun Ren, Jiliang Tang, and Dawei Yin.
\newblock Streaming graph neural networks.
\newblock In {\em Proceedings of the 43rd international ACM SIGIR conference on research and development in information retrieval}, pages 719--728, 2020.

\bibitem{mernyei2020wiki}
P{\'e}ter Mernyei and C{\u{a}}t{\u{a}}lina Cangea.
\newblock Wiki-cs: A wikipedia-based benchmark for graph neural networks.
\newblock {\em arXiv preprint arXiv:2007.02901}, 2020.

\bibitem{poursafaei2022towards}
Farimah Poursafaei, Shenyang Huang, Kellin Pelrine, and Reihaneh Rabbany.
\newblock Towards better evaluation for dynamic link prediction.
\newblock {\em Advances in Neural Information Processing Systems}, 35:32928--32941, 2022.

\bibitem{rossi2020temporal}
Emanuele Rossi, Ben Chamberlain, Fabrizio Frasca, Davide Eynard, Federico Monti, and Michael Bronstein.
\newblock Temporal graph networks for deep learning on dynamic graphs.
\newblock {\em ICML 2020 Workshop on Graph Representation Learning}, 2020.

\bibitem{roy2025llm}
Amit Roy, Ning Yan, and Masood Mortazavi.
\newblock Llm-driven knowledge distillation for dynamic text-attributed graphs.
\newblock {\em arXiv preprint arXiv:2502.10914}, 2025.

\bibitem{sen2008collective}
Prithviraj Sen, Galileo Namata, Mustafa Bilgic, Lise Getoor, Brian Galligher, and Tina Eliassi-Rad.
\newblock Collective classification in network data.
\newblock {\em AI magazine}, 29(3):93--93, 2008.

\bibitem{song2019session}
Weiping Song, Zhiping Xiao, Yifan Wang, Laurent Charlin, Ming Zhang, and Jian Tang.
\newblock Session-based social recommendation via dynamic graph attention networks.
\newblock In {\em Proceedings of the Twelfth ACM international conference on web search and data mining}, pages 555--563, 2019.

\bibitem{tang2023dynamic}
Haoran Tang, Shiqing Wu, Guandong Xu, and Qing Li.
\newblock Dynamic graph evolution learning for recommendation.
\newblock In {\em Proceedings of the 46th international acm sigir conference on research and development in information retrieval}, pages 1589--1598, 2023.

\bibitem{tang2024graphgpt}
Jiabin Tang, Yuhao Yang, Wei Wei, Lei Shi, Lixin Su, Suqi Cheng, Dawei Yin, and Chao Huang.
\newblock Graphgpt: Graph instruction tuning for large language models.
\newblock In {\em Proceedings of the 47th International ACM SIGIR Conference on Research and Development in Information Retrieval}, pages 491--500, 2024.

\bibitem{touvron2023llama}
Hugo Touvron, Louis Martin, Kevin Stone, Peter Albert, Amjad Almahairi, Yasmine Babaei, Nikolay Bashlykov, Soumya Batra, Prajjwal Bhargava, Shruti Bhosale, et~al.
\newblock Llama 2: Open foundation and fine-tuned chat models.
\newblock {\em arXiv preprint arXiv:2307.09288}, 2023.

\bibitem{trivedi2019dyrep}
Rakshit Trivedi, Mehrdad Farajtabar, Prasenjeet Biswal, and Hongyuan Zha.
\newblock Dyrep: Learning representations over dynamic graphs.
\newblock In {\em International conference on learning representations}, 2019.

\bibitem{velivckovic2018graph}
Petar Veli{\v{c}}kovi{\'c}, Guillem Cucurull, Arantxa Casanova, Adriana Romero, Pietro Li{\`o}, and Yoshua Bengio.
\newblock Graph attention networks.
\newblock In {\em International Conference on Learning Representations}, 2018.

\bibitem{wang2021tcl}
Lu~Wang, Xiaofu Chang, Shuang Li, Yunfei Chu, Hui Li, Wei Zhang, Xiaofeng He, Le~Song, Jingren Zhou, and Hongxia Yang.
\newblock Tcl: Transformer-based dynamic graph modelling via contrastive learning.
\newblock {\em arXiv preprint arXiv:2105.07944}, 2021.

\bibitem{wang2024beyond}
Xindi Wang, Mahsa Salmani, Parsa Omidi, Xiangyu Ren, Mehdi Rezagholizadeh, and Armaghan Eshaghi.
\newblock Beyond the limits: a survey of techniques to extend the context length in large language models.
\newblock In {\em Proceedings of the Thirty-Third International Joint Conference on Artificial Intelligence}, pages 8299--8307, 2024.

\bibitem{wang2021apan}
Xuhong Wang, Ding Lyu, Mengjian Li, Yang Xia, Qi~Yang, Xinwen Wang, Xinguang Wang, Ping Cui, Yupu Yang, Bowen Sun, et~al.
\newblock Apan: Asynchronous propagation attention network for real-time temporal graph embedding.
\newblock In {\em Proceedings of the 2021 international conference on management of data}, pages 2628--2638, 2021.

\bibitem{wang2021inductive}
Yanbang Wang, Yen-Yu Chang, Yunyu Liu, Jure Leskovec, and Pan Li.
\newblock Inductive representation learning in temporal networks via causal anonymous walks.
\newblock In {\em International Conference on Learning Representations (ICLR)}, 2021.

\bibitem{wang2024bridging}
Yaoke Wang, Yun Zhu, Wenqiao Zhang, Yueting Zhuang, Liyunfei Liyunfei, and Siliang Tang.
\newblock Bridging local details and global context in text-attributed graphs.
\newblock In {\em Proceedings of the 2024 Conference on Empirical Methods in Natural Language Processing}, pages 14830--14841, 2024.

\bibitem{wei2024llmrec}
Wei Wei, Xubin Ren, Jiabin Tang, Qinyong Wang, Lixin Su, Suqi Cheng, Junfeng Wang, Dawei Yin, and Chao Huang.
\newblock Llmrec: Large language models with graph augmentation for recommendation.
\newblock In {\em Proceedings of the 17th ACM International Conference on Web Search and Data Mining}, pages 806--815, 2024.

\bibitem{wu2019simplifying}
Felix Wu, Amauri Souza, Tianyi Zhang, Christopher Fifty, Tao Yu, and Kilian Weinberger.
\newblock Simplifying graph convolutional networks.
\newblock In {\em International conference on machine learning}, pages 6861--6871, 2019.

\bibitem{wu2024exploring}
Yuxia Wu, Shujie Li, Yuan Fang, and Chuan Shi.
\newblock Exploring the potential of large language models for heterophilic graphs.
\newblock {\em arXiv preprint arXiv:2408.14134}, 2024.

\bibitem{xu2020inductive}
Da~Xu, Chuanwei Ruan, Evren Korpeoglu, Sushant Kumar, and Kannan Achan.
\newblock Inductive representation learning on temporal graphs.
\newblock {\em International Conference on Learning Representations (ICLR)}, 2020.

\bibitem{xue2023efficient}
Rui Xue, Xipeng Shen, Ruozhou Yu, and Xiaorui Liu.
\newblock Efficient large language models fine-tuning on graphs.
\newblock {\em arXiv preprint arXiv:2312.04737}, 2023.

\bibitem{yan2023comprehensive}
Hao Yan, Chaozhuo Li, Ruosong Long, Chao Yan, Jianan Zhao, Wenwen Zhuang, Jun Yin, Peiyan Zhang, Weihao Han, Hao Sun, et~al.
\newblock A comprehensive study on text-attributed graphs: Benchmarking and rethinking.
\newblock {\em Advances in Neural Information Processing Systems}, 36:17238--17264, 2023.

\bibitem{yang2024qwen2}
An~Yang, Baosong Yang, Beichen Zhang, Binyuan Hui, Bo~Zheng, Bowen Yu, Chengyuan Li, Dayiheng Liu, Fei Huang, Haoran Wei, et~al.
\newblock Qwen2. 5 technical report.
\newblock {\em arXiv preprint arXiv:2412.15115}, 2024.

\bibitem{yang2021graphformers}
Junhan Yang, Zheng Liu, Shitao Xiao, Chaozhuo Li, Defu Lian, Sanjay Agrawal, Amit Singh, Guangzhong Sun, and Xing Xie.
\newblock Graphformers: Gnn-nested transformers for representation learning on textual graph.
\newblock {\em Advances in Neural Information Processing Systems}, 34:28798--28810, 2021.

\bibitem{yu2023towards}
Le~Yu, Leilei Sun, Bowen Du, and Weifeng Lv.
\newblock Towards better dynamic graph learning: New architecture and unified library.
\newblock {\em Advances in Neural Information Processing Systems}, 36:67686--67700, 2023.

\bibitem{zhangdtgb}
Jiasheng Zhang, Jialin Chen, Menglin Yang, Aosong Feng, Shuang Liang, Jie Shao, and Rex Ying.
\newblock Dtgb: A comprehensive benchmark for dynamic text-attributed graphs.
\newblock In {\em The Thirty-eight Conference on Neural Information Processing Systems Datasets and Benchmarks Track}, 2024.

\bibitem{zhang2022dynamic}
Mengqi Zhang, Shu Wu, Xueli Yu, Qiang Liu, and Liang Wang.
\newblock Dynamic graph neural networks for sequential recommendation.
\newblock {\em IEEE Transactions on Knowledge and Data Engineering}, 35(5):4741--4753, 2022.

\bibitem{zhang2025unifying}
Siwei Zhang, Yun Xiong, Yateng Tang, Xi~Chen, Zian Jia, Zehao Gu, Jiarong Xu, and Jiawei Zhang.
\newblock Unifying text semantics and graph structures for temporal text-attributed graphs with large language models.
\newblock {\em arXiv preprint arXiv:2503.14411}, 2025.

\bibitem{zhu2025graphclip}
Yun Zhu, Haizhou Shi, Xiaotang Wang, Yongchao Liu, Yaoke Wang, Boci Peng, Chuntao Hong, and Siliang Tang.
\newblock Graphclip: Enhancing transferability in graph foundation models for text-attributed graphs.
\newblock In {\em Proceedings of the ACM on Web Conference 2025}, pages 2183--2197, 2025.

\bibitem{zhu2024efficient}
Yun Zhu, Yaoke Wang, Haizhou Shi, and Siliang Tang.
\newblock Efficient tuning and inference for large language models on textual graphs.
\newblock In {\em Proceedings of the Thirty-Third International Joint Conference on Artificial Intelligence}, pages 5734--5742, 2024.

\end{thebibliography}

\medskip

\appendix
\section{Limitations}\label{sec:limitations}
One limitation of our paper is that we only test our method on DTGB benchmarks, considering that it is the only publicly available benchmark for DyTAGs. Experiments on more diverse real-world datasets and tasks will further demonstrate the usefulness of \MethodName. Another direction worthy of exploration is extending our method into more complicated graph types, such as DyTAGs with heterogeneous nodes and edges or hyper-relations.

\section{Notations}
In this section, we provide important notations used in this paper, detailed in Table \ref{tab:notation}.

\begin{table}[h]
    \centering
    \caption{Notations and descriptions.}
    \scalebox{0.8}{
    \begin{tabular}{l|l}
    \toprule
        \textbf{Notation}&\textbf{Description} \\
        \midrule
    $\mathcal{G}$ & A dynamic text-attributed graph. \\
    $\mathcal{E}$ & The edge set of the DyTAG. \\
    $\mathcal{T}$ & The timestamp set of the DyTAG. \\
    $\mathcal{D}$ & The node text attribute set of the DyTAG. \\
    $\mathcal{R}$ & The edge text attribute set of the DyTAG. \\
    $\mathcal{N}_v$ & The chronological list of interactions with node $v$'s first-hop neighbors.  \\
    $\mathcal{G}_t$ & The state of the DyTAG before timestamp $t$. \\
    $I_{u,v}$ & An interaction between two nodes. \\
    \midrule
    $\mathbf{F}^{\text{rc}}_i$ & The raw recent temporal semantic feature of an interaction $I_i$. \\
    $S$ & A segment of interactions, which contains interactions occurring between two partition timestamps. \\
    $\mathbf{T}^{\text{gb}}$ & The set of all partition timestamps. \\
    $D_i$ & The description of a node before a certain timestamp $S_i$. \\
     $\mathbf{F}^{\text{gb}}_i$ & The raw global temporal semantic feature of a node before the $i^{th}$ partition timestamp.\\
    \midrule
    $\mathbf{R}^{(l)}$ & The processed recent temporal semantic feature of the $l$-th layer.\\
    $\mathbf{G}^{(l)} $ & The processed global temporal semantic feature of the $l$-th layer.\\
    $\mathbf{S}^{(l)} $ & The graph structure feature given by a temporal GNN of the $l$-th layer \\
    $\mathcal{R}^r(\cdot)$ & A readout function to get the recent semantic feature of a certain node from a list of $\mathbf{R}^{(l)}$\\
    $\mathcal{R}^g(\cdot)$ & A readout function to get the global semantic feature of a certain node from a list of $\mathbf{G}^{(l)}$\\
        \bottomrule
    \end{tabular}}
    \label{tab:notation}
\end{table}

\section{Proof of Reasoning Complexity}
\label{sec:proof}
In this section, we give a complexity proof of recent semantic reasoning in Section \ref{sec:implicit reasoning} (i.e., the number of consumed input tokens), including both intuitive approach without node-centric reasoning and the proposed method with node-centric reasoning. For simplicity of analysis, we assume that the degree of each node is $d$ and the interaction of each node with its $d$ neighbor nodes occurs at distinct times $t_i (1 \leq i \leq d)$, where $t_i = t_j$ if and only if $i = j$. In this case, we have $O(|\mathcal{E}|) = O(|\mathcal{V}| \times d)$.

\textbf{Complexity of intuitive approach without node-centric reasoning.} Consider node $u$ and its historical interaction sequence:
\begin{equation}
    \mathcal{N}_u = \{I_{1}^u, I_{2}^u, ..., I_{d}^u\}, \text{where} \ t_1 \leq t_2 \leq ... \leq t_d.
    \label{eq:sequence}
\end{equation}
For a historical interaction $I_{i}^u$, assume its interacted node is $v$. To provide temporal context for $I_{i}^u$, the input content to the LLM is:
\begin{equation}
    \text{Input} = \{I_{j}^{k} | k \in \{u, v\}, 1 \leq j < i\} \cup \{I_{i}^u\}
\end{equation}
which totals $2\times(i-1) + 1$ interactions. Therefore, the input complexity for a single node is:
\begin{equation}
    O(\sum_{i=1}^{d} (2i-1)) = O(d^2).
\end{equation}
Since each edge belongs to two nodes, the total input complexity for all nodes is:
\begin{equation}
    O(\frac{|\mathcal{V}|}{2}) \times O(d^2) = O(|\mathcal{E}| \times d).
\end{equation}

\textbf{Complexity of our proposed method with node-centric reasoning.} Consider node $u$ and its historical interaction sequence $\mathcal{N}_u$, as shown in Eq. \ref{eq:sequence}. This sequence is divided by a sliding window into $\lceil\frac{d}{\frac{c}{2}}\rceil - 1$ input batches, where $c$ is the window length. The complexity of each batch is $O(c)$. Therefore, the input complexity for node $u$ is:
\begin{equation}
    O((\lceil\frac{d}{\frac{c}{2}}\rceil - 1) \times c) = O(d).
\end{equation}
The final input complexity for all nodes is $O(d \times |\mathcal{V}|) = O(|\mathcal{E}|)$.

\section{Prompt for LLM Reasoning}
In this section, we provide dataset-specific prompt templates used for LLM reasoning, including implicit reasoning for recent temporal semantics and explicit reasoning for global temporal semantics. We select prompt for Googlemap as an example. The remaining dataset follows a similar structure.

\begin{figure}[!ht]
\centering
\includegraphics[width=1.0\columnwidth]{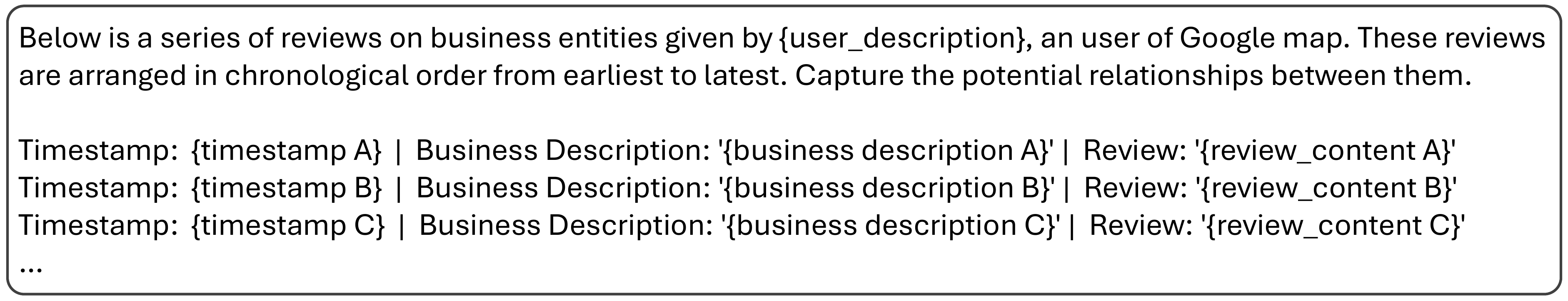}
\caption{The prompt used for Googlemap of recent semantic reasoning. }
\label{fig:recent_prompt}
\end{figure}

\begin{figure}[!ht]
\centering
\includegraphics[width=1.0\columnwidth]{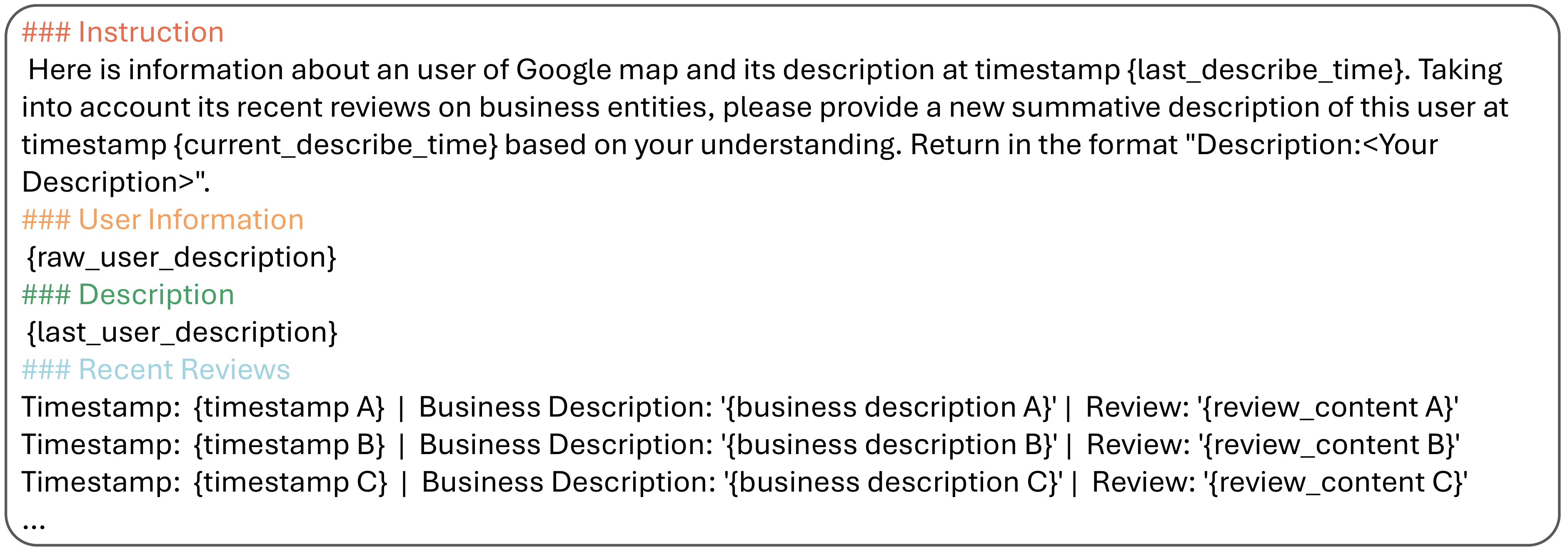}
\caption{The prompt used for Googlemap of global semantic reasoning.}
\label{fig:global_prompt}
\end{figure}

\section{Detailed Experimental Settings}\label{sec:detail_setting}

\textbf{Evaluation Tasks and Metrics.} As detailed in Section \ref{sec:notations}, following \citep{zhangdtgb}, we evaluate models using the destination node retrieval task under two settings: the transductive setting where the model predicts future links between nodes observed during training, and the inductive setting where the model predicts links between unseen nodes. We use the Hit@k as evaluation metrics, which represents the proportion of test instances where the score of the interaction between the source and ground-truth destination node ranks within the top-k among scores for all candidate nodes. Additionally, we present results on the future link prediction task for reference with AP and ROC-AUC metrics.

\textbf{Model Configurations.} For \MethodName, we primarily utilize DyGFormer as the base temporal GNN and include TGAT to validate the generalizability. We mainly employ Llama-3.1-8B / Llama-3.1-8B-Instruct for implicit and explicit reasoning, respectively. We also include Qwen2.5-7B / Qwen2.5-7B-Instruct \citep{yang2024qwen2} and Mistral-7B / Mistral-7B-Instruct \citep{DBLP:journals/corr/abs-2310-06825} to verify the generalizability of \MethodName~across different LLMs. We set $c$ to 64 for GDELT and 16 for the remaining datasets. $c$ is selected based on following considerations:
\begin{itemize}[leftmargin=15pt, itemsep=0pt]
\item \textbf{Data Characteristics:} $c$ is adapted to the average node degree. Datasets with denser histories, like GDELT, require a larger $c$ to capture sufficient temporal context.
\item \textbf{Computational \& LLM Constraints:} A larger $c$ increases input sequence length, posing challenges for computational resources and the LLM's hard token limit. For datasets with lengthy text attributes (e.g., Enron and Stack\_elec), $c$ is carefully bounded by these resource constraints.
\item \textbf{Model Performance:} Crucially, LLM performance degrades on excessively long inputs, especially as they approach or exceed the model's context window—a phenomenon we analyze in Figure \ref{fig:sensitive}. This is the most critical interaction with token limits. We select $c$ to maximize information gain while avoiding this performance cliff.
\end{itemize}
We set $s$ to 8 across all datasets. The impact of these two hyperparameters is investigated in Section \ref{sec:ablation}. The total number of layers $L$ is set to 2.

\textbf{Implementation Details.} We adopt the experimental setup provided by DTGB benchmark. All datasets are split into training, validation, and test sets with a ratio of 0.7:0.15:0.15. All models are trained using the Adam \citep{kingma2014adam} optimizer with a batch size of 256 and a learning rate of 0.0001. Training runs for up to 50 epochs, with validation performed every 5 epochs. We employ an early stopping mechanism with a patience of 5 epochs. Model-specific hyperparameters for each baseline are set according to the optimal values recommended by the DTGB benchmark (see Table \ref{tab:baseline_setting} for details). Text attributes within the datasets, as well as the text generated by the LLM during explicit reasoning, are vectorized with the bert-base-uncased model \citep{devlin2019bert}. All training was conducted on NVIDIA A100 40G GPUs.

\textbf{Training for \MethodName.} Following DTGB, we supervise the model training using a binary cross-entropy loss. Specifically, we use an MLP layer to take the final features (i.e., $\mathbf{M}^{(L)}_u$ and $\mathbf{M}^{(L)}_v$) to predict the probability score of a future interaction between nodes $u$ and $v$. For each true interaction $I_{u,v_{pos}}$ occurring at time $T$, we randomly sample a negative node $v_{neg}$ to create a false interaction $I_{u,v_{neg}}$, which did not actually occur. Finally, we use all the true and false samples for training.

\subsection{Descriptions of Datasets}
\label{sec:description_datasets}
\begin{figure}[!ht]
\centering
\includegraphics[width=0.8\columnwidth]{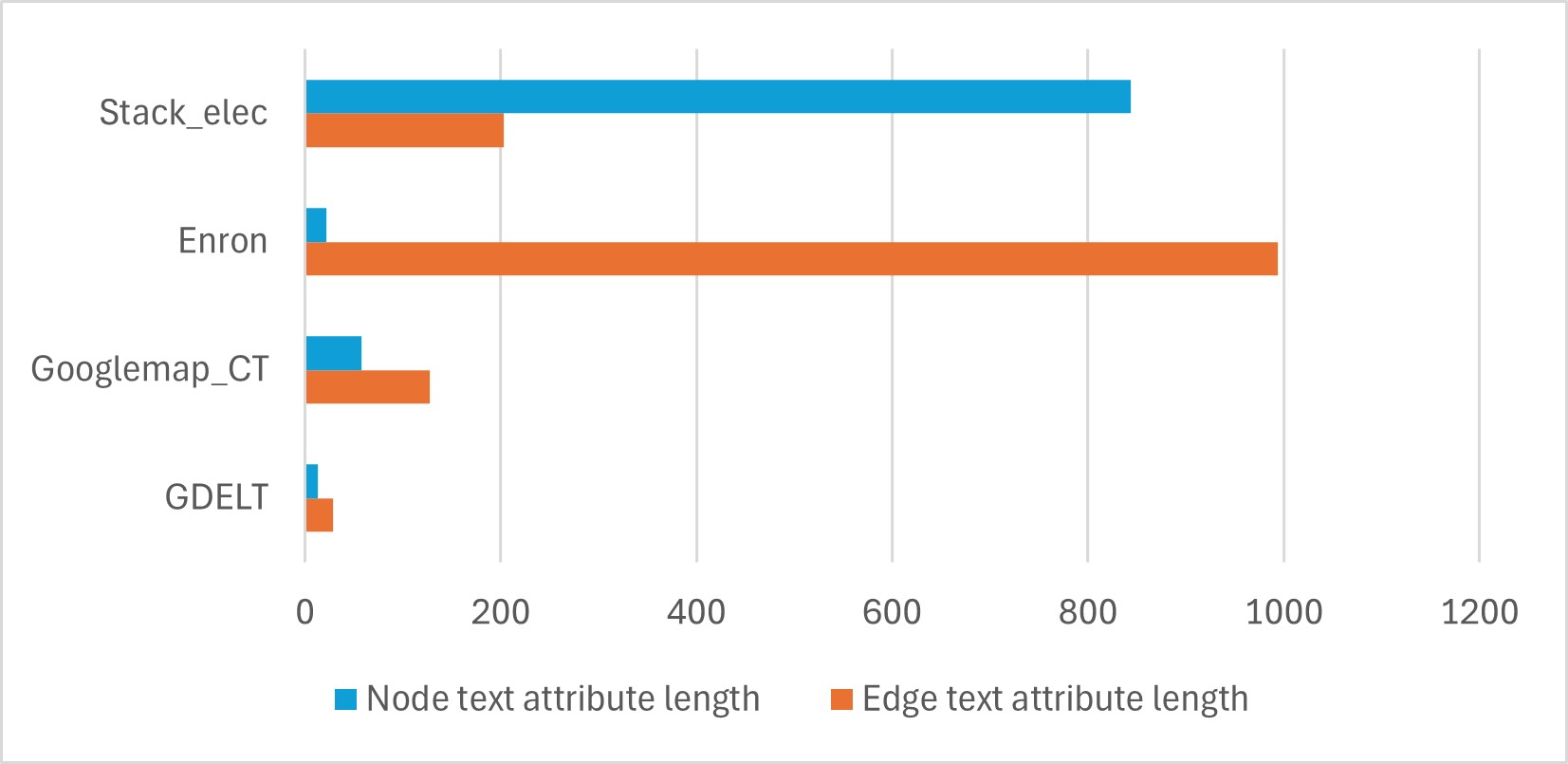}
\caption{Text attribute length of used DyTAG datasets.}
\label{fig:dataset_text}
\end{figure}

We use four datasets for the experiments, including GDELT, Enron, Googlemap and Stack\_elec. We provide the statistics of datasets in Table \ref{tab:dataset_data}. We also summarize the text attribute length of these datasets in Figure \ref{fig:dataset_text}.

\begin{table*}[h]
    \centering
    \caption{Statistics of datasets.}
    \scalebox{1}{
    \begin{tabular}{l|ccccc}
    \toprule
    \textbf{Dataset}&\textbf{Nodes}&\textbf{Edges}&\textbf{Domain}&\textbf{Timestamps} & \textbf{Bipartite Graph}\\
        \midrule
        \textbf{GDELT} & 6,786 & 1,339,245 & Knowledge graph & 2,591 & $\times$ \\
        \textbf{Enron} & 42,711 & 797,907 & E-mail & 1,006 & $\times$ \\
        \textbf{Googlemap} & 111,168 & 1,380,623 & E-commerce & 55,521 & $\checkmark$ \\
        \textbf{Stack\_elec} & 397,702 & 1,262,225 & Multi-round dialogue & 5,224 & $\checkmark$ \\
        \bottomrule
    \end{tabular}}
    \label{tab:dataset_data}
\end{table*}

We provide descriptions of these datasets as follows.
\begin{itemize}[leftmargin=0.5cm]
    \item \textbf{GDELT}: this dataset comes from the "Global Database of Events, Language, and Tone" project, which aims to build a catalog of political behavior covering countries worldwide. In this dataset, nodes represent political entities and the node text attribute is the entity name while edges represent political relationships or actions between entities and the edge text attribute represents the description of the behavior.
    \item \textbf{Enron}: this dataset originates from the email communications of employees at the Enron energy company. Here, nodes represent employees and the node text attribute is the employee's email address while edges represent the sending of emails and the edge text attribute is the email content.
    \item \textbf{Googlemap}: this dataset is extracted from the Connecticut State (CT) portion of the "Google Local Data" project, containing business review information from Google Maps. In this dataset, nodes are users or businesses and the node text attribute is the user name or business description while edges represent reviews from users to businesses and the edge text attribute is the specific review content.
    \item \textbf{Stack\_elec}: this dataset comes from anonymized data of user question-and-answer content related to electronics technology on the Stack Exchange website. In this dataset, nodes are questions or users on the site and the node text attribute is the question description or user profile while edges represent answers or comments by users on questions and the edge text attribute is the content of the answer or comment.
\end{itemize}

\subsection{Description of baselines}
\label{sec:description_baselines}
In this section, we provide detailed descriptions of the baselines.  
\begin{itemize}[leftmargin = 0.5cm]
    \item \textbf{JODIE}~\citep{kumar2019predicting}: JODIE is a coupled recurrent neural network model that learns dynamic embedding trajectories for users and items based on their temporal interactions. JODIE utilizes a novel projection operator to predict the future trajectory of embeddings and employs a scalable t-Batch training algorithm, significantly outperforming baseline methods in future interaction and user state change prediction tasks.
    \item \textbf{DyRep}~\citep{trivedi2019dyrep}: DyRep is a framework for learning representations on dynamic graphs by modeling topological evolution and node interactions as two distinct but coupled temporal point processes. It uses a temporal-attentive network to learn evolving node embeddings that capture the interplay between these structural and activity dynamics over continuous time.
    \item \textbf{CAWN}~\citep{wang2021inductive}: CAWN is a neural network model for inductive representation learning on temporal networks. CAWN captures network dynamics using temporal random walks and a novel anonymization strategy, outperforming prior methods in predicting future links, particularly on unseen parts of networks.
    \item  \textbf{TCL}~\citep{wang2021tcl}: TCL is a Transformer-based dynamic graph modeling method, which optimizes dynamic node representations through contrastive learning. TCL adopts a two-stream encoder architecture to respectively process the temporal neighborhood information of the target interaction nodes and fuses them at the semantic level via a co-attentional Transformer.
    \item \textbf{GraphMixer}~\citep{cong2023we}: GraphMixer is a conceptually simple architecture for temporal link prediction using only MLPs and neighbor mean-pooling, intentionally avoiding complex RNN and self-attention mechanisms. Despite its simplicity, GraphMixer achieves state-of-the-art performance on benchmarks, demonstrating better results, faster convergence, and improved generalization compared to more complicated models.
    \item \textbf{TGAT}~\citep{xu2020inductive}: TGAT utilizes a self-attention mechanism combined with a novel functional time encoding technique based on harmonic analysis. TGAT efficiently aggregates temporal-topological neighborhood information to generate time-aware embeddings for both existing and newly appearing nodes.
    \item \textbf{DyGFormer}~\citep{yu2023towards}: DyGFormer is a new Transformer-based architecture for dynamic graph learning that uses neighbor co-occurrence encoding and sequence patching to effectively capture node correlations and long-term dependencies from historical first-hop interactions.
\end{itemize}

\subsection{Configurations of Different Methods}
\label{sec:baseline_setting}
We present the hyperparameters for all baselines in Table \ref{tab:baseline_setting}, which are recommended by the official DTGB benchmark.
\begin{table*}[h]
    \centering
    \caption{Hyper-parameter setting for all baselines.}
    \resizebox{\textwidth}{!}{
    \begin{tabular}{lccccccc}
    \toprule
    \textbf{Methods}&\textbf{\# layers}&\textbf{\# heads}&\textbf{dropout}&\textbf{time\_feat\_dim}&\textbf{channel\_embedding\_dim}&\textbf{patch\_size}&\textbf{max\_input\_sequence\_length} \\
        \midrule
        JODIE & 1 & 2 & 0.1 & 100 & / & / & / \\
        DyRep & 1 & 2 & 0.1 & 100 & / & / & / \\
        CAWN & / & / & 0.1 & 100 & / & / & / \\
        TCL & 2 & 2 & 0.1 & 100 & / & / & /  \\
        GraphMixer & 2 & / & 0.1 & 100 & / & / & /  \\
        TGAT & 2 & 2 & 0.1 & 100 & / & / & /  \\
        DyGFormer & 2 & 2 & 0.1 & 100 & 50 & 1 & 48 \\
    \midrule
        \textbf{Methods}&\textbf{\# depths}&\textbf{\# walk\_heads}&\textbf{walk\_length}&\textbf{position\_feat\_dim}&\textbf{\# neighbors}&\textbf{sample\_neighbor\_strategy}&\textbf{time\_scaling\_factor} \\
        \midrule
        JODIE & / & / & / & / & 10 & recent & / \\
        DyRep & / & / & / & / & 10 & recent & / \\
        CAWN & / & 8 & 1 & 768 & 32 & time\_interval\_aware & 1e-6 \\
        TCL & 21 & / & / & / & 20 & recent & /  \\
        GraphMixer & / & / & / & / & 20 & recent & /  \\
        TGAT & / & / & / & / & 20 & recent & /  \\
        DyGFormer & / & / & / & / & / & recent & / \\
        \bottomrule
    \end{tabular}
    }
    \label{tab:baseline_setting}
\end{table*}

\end{document}